\documentclass[trsc,nonblindrev]{informs3} 

\OneAndAHalfSpacedXI 

\usepackage{natbib}
\usepackage[T1]{fontenc}
 \bibpunct[, ]{(}{)}{,}{a}{}{,}%
 \def\bibfont{\small}%
\usepackage{tabularx}
\usepackage{booktabs}
\usepackage{multirow}
\usepackage{makecell}
\usepackage{algorithm}
\usepackage{algpseudocode}
\usepackage{threeparttable}

\TheoremsNumberedThrough     

\EquationsNumberedThrough    

\begin{document}



\RUNTITLE{Decision Aware OD Estimation}

\TITLE{Decision Aware Origin-Destination Demand Estimation with Structurally Constrained Data}


\ARTICLEAUTHORS{
\AUTHOR{Xianzhe Wang}
\AFF{Institute of Transportation Studies, University of California, Davis, CA, USA, \EMAIL{xzhwang@ucdavis.edu}}
\AUTHOR{Yueyue Fan}
\AFF{Department of Civil and Environmental Engineering, University of California, Davis, CA, USA, \EMAIL{yyfan@ucdavis.edu (corresponding author)}}
\AUTHOR{Johannes O. Royset}
\AFF{Daniel J. Epstein Department of Industrial and Systems Engineering, University of Southern California, CA, USA, \EMAIL{royset@usc.edu }}
} 

\ABSTRACT{%
Origin-destination (OD) demand plays an important role in transportation system design and management applications. Conventional OD estimation methods typically judge estimation quality based on how estimated demand aligns with observed traffic data and historical demand patterns. We argue that data alignment alone is insufficient; how errors in OD estimates influence downstream decision quality should also be considered. This paper develops a decision aware OD demand estimation framework that leverages structurally constrained traffic data to characterize an admissible demand region, within which we identify OD estimates that would lead to least regret in the downstream decision. Therefore, our OD estimates are both data-consistent and decision-relevant.   Methodologically, we approximate the admissible demand region induced by physical network constraints through a geometric representation, which regularizes the search over feasible OD estimates. Within this regularized region, we use a convex surrogate of downstream regret to obtain a tractable decision aware optimization problem. Numerical experiments show that the proposed framework consistently reduces decision regret relative to conventional estimation benchmarks, demonstrating the value of aligning OD demand estimation with downstream decision quality.
}%


\HISTORY{}

\maketitle

%


\section{Introduction}

Travel demand serves as a critical parameter in almost all transportation system design and management applications, including traffic assignment, network design, congestion pricing, and infrastructure investment \citep{patriksson2015traffic,yang1998models,de2011traffic,rokicki2018major}.  In the transportation science literature, the characterization and quantification of travel demand have long been central research objectives.  Much of the existing literature on demand estimation focuses primarily on data alignment, where success is measured by how well estimated origin-destination (OD) flows conform to observed measurements or long-run covariates \citep{bierlaire2004efficient, krishnakumari2019data, demissie2022estimation}. In most transportation applications, OD demands are not ends in themselves, but rather intermediate inputs to downstream optimization, control, or policy design problems.  Despite the central role of travel demand played in downstream applications, established methods for demand estimation remain decision agnostic.  In this paper, we begin from the premise that the value of travel demand estimates lies in the quality of the decisions they support, not merely in their fidelity to observed data. Building on this premise, we introduce a new modeling framework for estimating the OD demand that is consistent with observed traffic conditions as well as its impact on the downstream decision-making objectives. 


Existing travel demand estimation methods can be broadly categorized into two complementary paradigms: demand forecasting and demand reconstruction.  Demand forecasting is for longer-term prediction and is typically grounded in the modeling of socioeconomic factors, such as household characteristics, economic activities, land use and accessibility. that give rise to travel demand \citep{jin2014location, pourebrahim2019trip, yao2020spatial, yin2023convgcn}. 
Demand reconstruction typically aims to infer an OD demand that best explains observed traffic conditions, thus often explicitly builds on network-based models.  In this regard, our study fits better within the category of demand reconstruction, though our estimation framework leaves some flexibility to incorporate predictions from established behavioral models as prior information.  


A major challenge in demand reconstruction lies in the underdetermined nature of the inverse network problem \citep{wang2023transportation}. Data consistency alone provides an incomplete basis for evaluating OD estimates, as multiple demand patterns may be equally compatible with observed measurements. A common approach to address the non-uniqueness of the inferred demand is to incorporate prior OD information, either for the complete set of OD pairs in a network \citep{cascetta1984estimation, castillo2008observability} or for a partially observed subset of OD demands \citep{yang2015data}. More broadly, uniqueness has been achieved by embedding additional structural preferences into the estimation problem, such as entropy maximization \citep{xie2011maximum, Bar-Gera2006RouteFlow}, sparsity regularization \citep{wang2023transportation}, and statistical or dynamical assumptions \citep{maher1983inferences, ros2022practical}, which effectively shrink the feasible OD set and select a single admissible solution consistent with observed traffic data. All of these approaches are driven primarily by data alignment, with little consideration given to the quality of the downstream decisions that the resulting OD estimates are intended to support.


This paper presents a decision aware framework for OD estimation, shifting the focus from aligning estimates with data to aligning them with the performance of downstream decision-making tasks. To operationalize this decision awareness, we introduce decision regret as a bridge between OD estimation and downstream decision-making. Regret quantifies the loss in downstream objective value, had the estimated demand missed the ``true'' value.  The concept of regret has been widely adopted in operations research as a principled criterion for assessing decision quality, particularly in evaluating the robustness and economic consequences of imperfect information \citep{bell1983risk, donti2017task, elmachtoub2022smart}.

We consider a downstream optimization problem that depends on the OD demand $f$. Let $\mathcal{F}$ denote the feasible OD demand space, consisting of OD demand vectors that are consistent with the observed network data and satisfy basic feasibility constraints, such as nonnegativity and prescribed lower and upper bounds on OD demand. For any $f \in \mathcal{F}$, let $w(f) \in \arg\min_{w \in \mathcal{W}} f^\top w$ denote an optimal downstream decision. Given the true OD demand $f_{\rm true}$, the decision regret of an estimated demand $f$ is defined as
\begin{equation}\label{eq-regret}
    R(f_{\rm true}, f)
=
{f_{\rm true}}^{\top} w(f)
-
\min_{w \in \mathcal{W}} {f_{\rm true}}^{\top} w.
\end{equation}
Since $f_{\rm true}$ is unobserved in practice, we use the available prior demand as a proxy for $f_{\rm true}$ when constructing the regret criterion. The resulting formulation combines three key elements: link flow observations and feasibility constraints define the admissible OD space $\mathcal{F}$, the prior demand provides the reference for regret evaluation, and the downstream decision objective measures the performance of candidate OD estimates.




This research design departs from classical OD estimation in three ways. First, it replaces data fit as the primary criterion with downstream decision performance, directly linking OD estimation with downstream optimization. Second, it recognizes that OD estimation errors differ in their decision impact, giving greater weight to errors that are more consequential for downstream decisions. Third, this perspective treats the non-uniqueness of OD reconstruction not as a deficiency, but as an opportunity to select among feasible OD solutions based on their downstream decision performance.

Certainly, the new estimation framework also raises several challenges. The feasible OD demand space $\mathcal{F}$ induced by structural constraints on the data can be high-dimensional and geometrically complex, while the regret function is generally nonconvex and thus computationally challenging to optimize. Moreover, directly optimizing a linear objective over $\mathcal{F}$ tends to yield an extreme-point solution, which may be overly sensitive to sample-specific noise and thereby increase the risk of overfitting. To address these challenges, we develop a tractable decision aware estimation framework that uses geometric approximations of the feasible OD space to regularize estimation and mitigate sensitivity to extreme-point solutions, and replaces the nonconvex regret function with a convex surrogate. Exploiting duality, we reformulate the resulting problem as a convex quadratically constrained program, enabling scalable identification of structurally admissible OD estimates that support high-quality downstream decisions.


The main contributions of this study are as follows. From a conceptual perspective, we formally introduce a new problem formulation that systematically integrates structurally constrained OD reconstruction and downstream decision evaluation within a unified optimization framework. From a computational perspective, we develop a tractable and scalable decision aware estimation framework. The proposed approach comprises two key components: (i) a geometric ellipsoidal approximation of the structurally admissible OD demand space to address underdetermination and mitigate overfitting; (ii) a convex surrogate of the regret objective that preserves sensitivity to downstream decisions. From an empirical perspective, we develop a computationally efficient and scalable solution method and conduct extensive numerical experiments to validate the proposed framework. The results demonstrate that the decision aware OD estimates consistently yield lower regret in downstream decision problems compared with conventional methods, highlighting the practical value of aligning OD estimation objectives with decision performance.

The remainder of the paper is organized as follows. Section~\ref{DAODE-LIT} summarizes the current state of knowledge in the broader decision aware estimation literature, and identifies remaining gaps arising from unique challenges of transportation applications. Section~\ref{DAODE-FRAM} develops the decision aware OD estimation framework. It constructs a geometric approximation of the admissible OD demand space from structurally constrained data, formulates the corresponding decision aware estimation problem, and develops surrogate regret formulations that recast this problem as a tractable convex program. Section~\ref{DAODE-EXP} presents numerical experiments on a range of network settings to evaluate the proposed approach in terms of downstream decision performance. Section~\ref{DAODE-CON} concludes with a discussion of implications and directions for future research.

\section{Relation to the Broader Literature on Decision Aware Estimation}\label{DAODE-LIT}

The use of downstream decision models, typically formulated as optimization problems, to inform estimation and learning has a long conceptual history and has attracted renewed attention in recent decades. As early as 1983, \cite{bell1983risk} introduces decision regret as a decision level concept based on counterfactual comparisons, laying the conceptual foundation for later decision-based evaluation frameworks, though not explicitly for estimation or prediction. \cite{bengio1997using} articulates a decision centric learning paradigm, arguing that when downstream decision performance is the ultimate objective, minimizing prediction error is inherently suboptimal. Building on this insight, \cite{kao2009directed} proposes directed regression, which explicitly embeds directional sensitivity induced by downstream objectives into the estimation process, thereby shifting the evaluation of estimation from standalone predictive accuracy to its consequences for downstream decision quality. Subsequent studies have shown the practical relevance of decision aware estimation across diverse domains, including operational planning, capacity management, and inventory control \citep{liyanage2005practical, bassamboo2009data, akcay2011improved, li2026data}, consistently finding that gains in predictive accuracy do not necessarily translate into improved decision quality.

In recent years, there has been a surge of interest in systematically incorporating downstream decision models into estimation and learning procedures. A canonical example is the Smart "Predict, then Optimize" (SPO) framework introduced by \cite{elmachtoub2022smart}. For optimization problems with uncertain linear objective coefficients, the framework introduces the SPO loss, which uses decision regret rather than traditional prediction losses (e.g., mean squared error) to penalize the downstream optimality gap, and admits a convex surrogate known as the SPO+ loss. In a sequence of subsequent studies, \cite{elbalghiti2023generalization} establishes finite-sample generalization bounds for the SPO loss function. Building on this line of work, \cite{loke2022decision} proposes a decision-driven regularization framework that provides a unifying perspective on balancing prediction accuracy and decision quality, with the SPO+ loss arising as a special case under specific choices of the regularization parameters. \cite{mandi2020smart} demonstrates the empirical effectiveness of the SPO+ loss across a range of combinatorial optimization problems. Beyond seeking surrogate losses for decision regret, a broad stream of work aims to directly approximate the sensitivity of downstream decisions with respect to predicted parameters. Specifically, this line of research focuses on estimating the gradient $\frac{\partial w}{\partial f}$ and leveraging this information to capture how changes in the predictive model affect downstream decisions. The resulting sensitivity information is then used to guide the parameter updates of the estimation model. Some studies restrict attention to structured classes of optimization problems. For example, \cite{amos2017optnet} consider convex quadratic programs and obtain analytical expressions for the decision gradients by differentiating through the Karush–Kuhn–Tucker (KKT) conditions via the implicit function theorem. Related approaches have been developed for conic optimization problems \citep{agrawal2019differentiating}. However, for general optimization problems, this gradient-based paradigm faces a fundamental challenge because the mapping from problem parameters to optimal decisions can be nonsmooth or unstable; see, for example, Section 5.A of \cite{royset2021optimization} for a discussion of the instability of minimizers. In many discrete optimization problems, this mapping is piecewise constant, implying that the corresponding derivatives are zero almost everywhere and undefined at the remaining points. To address this issue, existing approaches introduce regularization to construct a surrogate for the decision gradient \citep{wilder2019melding, mandi2020interior}. Beyond explicitly regularizing the optimization objective, several studies adopt a probabilistic perspective and introduce implicit regularization through randomized decision mappings. In this line of work, the mapping from predictions $f$ to decisions $w$ is modeled as a function that induces a probability distribution over the feasible region $\mathcal W$, so that $w(f)$ can be interpreted as a random variable conditional on $f$. This formulation enables gradient estimation via perturbation-based sampling schemes \citep{poganvcic2019differentiation,blondel2020learning}. A substantial body of subsequent work has examined the statistical properties, surrogate loss constructions, and computational tractability of decision aware objectives, with demonstrated applicability to a broad class of downstream decision problems. We refer the reader to recent survey papers in this area for a comprehensive overview \citep{sadana2025survey, mandi2024decision, qi2022integrating}.

\paragraph{Limitations and gaps in transportation applications.}
While these paradigms have substantially advanced the integration of estimation and optimization, they remain fundamentally misaligned with the structure of transportation network-level OD demand estimation problems. Here, OD demand estimation is characterized by high-dimensional feasible demand sets induced by physical network constraints, flow conservation, and behavioral assumptions, together with limited and indirect observations. In such settings, OD demand is generically non-identifiable: many distinct demand patterns are observationally equivalent, yet can give rise to sharply different downstream decisions. Existing decision aware approaches, however, largely retain a point estimation perspective, implicitly framing estimation as the recovery of a single latent ``true” parameter vector and interpreting decision losses as the consequence of small, local deviations around this point. This viewpoint is ill-suited to OD estimation, where uncertainty is not concentrated around a neighborhood of a unique solution, but is distributed over a large, structured feasible region.  The proposed framework integrates structurally constrained OD estimation with decision aware optimization to fill the gaps. 



\section{Decision Aware OD Estimation Framework}\label{DAODE-FRAM}

This section develops a decision aware framework for OD demand estimation in transportation networks. In Section~\ref{FRAM-E}, we characterize the structurally underdetermined OD estimation problem induced by network flow constraints and construct a tractable ellipsoidal approximation of the admissible OD space. Building on this geometric representation, in Section~\ref{Framework}, we then formulate a decision aware OD estimation framework that explicitly incorporates downstream optimization problems and defines estimation objectives in terms of expected downstream decision regret under a given error distribution of the true demand. Section~\ref{SOL-SPO} introduces a convex surrogate for the generally nonconvex regret objective, yielding tractable surrogate regret formulations. Finally, Section~\ref{sec:computational_formulation} derives deterministic equivalent reformulations of the resulting models, showing that the formulation over the original feasible OD region reduces to a linear program, whereas its ellipsoidal counterpart admits an equivalent convex quadratically constrained programming formulation.

\subsection{Geometric Approximation from Structurally Constrained Data}\label{FRAM-E}

A defining feature of transportation systems is that many observed quantities are governed by physical conservation laws imposed by the network structure. In particular, when traffic moves through a fixed set of links and originates from a collection of OD pairs, the relationship between link flows and OD flows is fully encoded by the \emph{link-OD incidence matrix}. Let $X\in\mathbb{R}^{m\times n}$ denote this matrix, where $m$ is the number of directed links and $n$ is the number of OD pairs. Given a vector of reliable link-flow measurements $L\in\mathbb{R}^m$, the underlying OD flow vector $f\in\mathbb{R}^n$ must satisfy the fundamental physical constraint
\begin{equation}
    L = X f. 
    \label{eq:flow_conservation}
\end{equation}
Equation~\eqref{eq:flow_conservation} captures the conservation of vehicles on every link of the network: the total number of vehicles that traverse a link must equal the aggregate contributions of all OD pairs that utilize that link.

For realistic networks, the number of OD pairs far exceeds the number of links, i.e., $n \gg m$. Consequently, the system of equations in \eqref{eq:flow_conservation} is \emph{underdetermined}. For any observed link-flow vector $L$, there exist infinitely many estimators for $f$, but they must respect the feasible set defined by \eqref{eq:flow_conservation}. Put differently, even though the OD flows are unobserved and often highly uncertain, they are never arbitrary: they are restricted to lie within the affine subspace of $\mathbb{R}^n$ determined by the measured link flows and the physical network structure.


For a linear system $Xf=L$ where $X$ is rank deficient (\,$n \gg m$\,), we highlight two remarks that are directly relevant to this study.  First, when $X^\top X$ is nonsingular (the full rank case), Moore-Penrose pseudoinverse $X^+$ is equivalent to $X^{+} = (X^\top X)^{-1}X^\top$, where $X^{-1}$ is the ordinary inverse of matrix $X$.  Second, the solution $X^+L$, calculated based on the Moore-Penrose pseudoinverse, has the minimum length among all feasible solutions to  $Xf=L$ (see \cite{FAN2003323Peudoinverse, rao1971generalized} for more comprehensive discussions). Within the broader feasible OD demand space $\mathcal{F}$, we consider the subset $S$ characterized by consistency with the observed link flows. With this notation, the full solution set to the normal equation is
\begin{equation}
    S
    = X^{+} L + \ker(X^\top X)
    = (X^+ L 
      \;+\; \{\, v \in \mathbb{R}^n : X^\top X v = 0 \,\}.
    \label{eq:solution_set}
\end{equation}
This decomposition reveals that all feasible OD flow vectors consist of a canonical
minimum norm solution plus an arbitrary vector in the nullspace of $X^\top X$, 
which captures all OD flow directions that do not affect the observed link flows.

Let $q=\operatorname{rank}(X)$ and let $\bar f = X^{+}L$ denote the particular least squares solution. We express the feasible set for the demand estimator as
$$
    S' = \{\;\bar f + V z \;\mid\; 0 \le \bar f + V z \le U, z\in \mathbb{R}^{n-q} \;\},
$$
where $V\in\mathbb{R}^{n\times(n-q)}$ is the matrix whose columns form an orthonormal basis for
$\ker(X^\top X)$, and $U$ sets an upper bound. This representation follows directly from the fact that the nullspace of
$X^\top X$ contains all directions that leave the link flows unchanged; hence,
any feasible $f$ differs from $\bar f$ only through motions in this subspace.

In a meaningful size of network problem, the feasible region $S'$ constitutes a high-dimensional polyhedral set. Because direct characterization of its size and shape is computationally intractable, we approximate this feasible region by an ellipsoid, which provides a tractable representation of its geometric extent and uncertainty.  Following the approach of
\cite{renegar2001mathematical} and \cite{chen2006confidence}, we construct an ellipsoidal
approximation centered at the analytic-center of $S'$.  
Let $z^\star$ denote the analytic-center parameter, defined by the solution to
the logarithmic barrier problem
$$
    \min_{z} \;
    -\sum_{i=1}^n 
    \Bigl[
         \log(U_i - \bar f_i - (Vz)_i)
         \;+\;
         \log(\bar f_i + (Vz)_i)
    \Bigr].
$$
For notational simplicity, define the vector $
    g(z) = \bar f + V z$, 
so that the objective becomes $ \phi(z) 
    = 
    -\sum_{i=1}^n 
    \Bigl[
        \log(U_i - g_i(z)) 
        + 
        \log(g_i(z))
    \Bigr]$. 

Let $H$ be the Hessian of $\phi(z)$ at $z^\star$.  
Since $g(z) = \bar f + Vz$ is affine, its Jacobian is $V$, and the Hessian takes 
the explicit form
$$
H=V^\top \left[\mathrm{diag}\!\left(\frac{1}{g_i(z^\star)^2}+\frac{1}{(U_i - g_i(z^\star))^2}\right)\right]V.
$$
This Hessian defines the local curvature of the logarithmic barrier and yields
a natural ellipsoidal approximation of $S'$ centered at the analytic-center
$f^\star = g(z^\star)$. Following Renegar~(2001), define the $H$-norm ball $B(z^\star, r) = \bigl\{z \;\bigm|\; (z - z^\star)^\top H (z - z^\star) \le r^2 \bigr\}$. Renegar's analysis establishes that, for self-concordant logarithmic barriers,
$$
    B(z^\star, 1)
    \;\subseteq\;
    \{\,z \mid 0 \le \bar f + V z \le U\,\}
    \;\subseteq\;
    B(z^\star, 4\nu + 1),
$$
where $\nu$ is the complexity parameter of the barrier function. In our case, the barrier consists of $2n$ logarithmic terms, and therefore $\nu = 2n$.

\begin{figure}[t]
    \centering
    \includegraphics[width=0.75\linewidth]{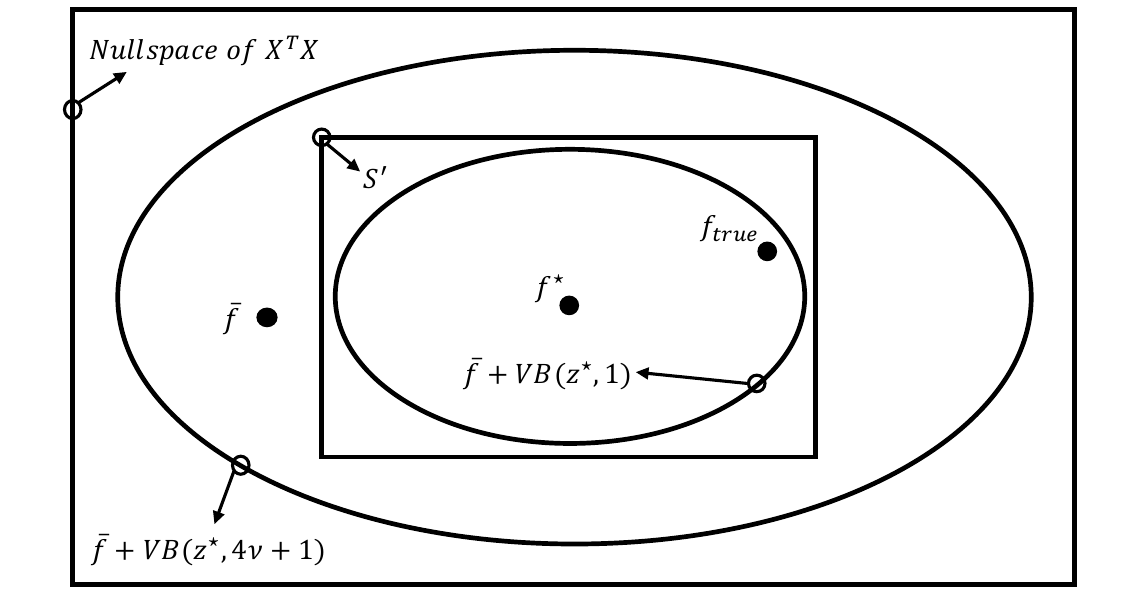}  
    \caption{
    Schematic representation of the geometric structure induced by the 
    link-OD incidence matrix.  
    The outer region depicts the nullspace of $X^TX$. The rectangular region represents the bounded feasible set 
    $S' = S \cap [0,U]$. The ellipses illustrate the inscribed and circumscribed ellipsoidal 
    approximations $\bar f + V B(z^\star,1)$ and $\bar f + V B(z^\star,4\nu+1)$. The figure also highlights the true OD flow $f_{\rm true}$, the least squares estimate $\bar f$, and the analytic-center estimate $f^{\star}$.
    }
    \label{fig:geometry}
\end{figure}

Mapping these sets back to the OD flow space via the affine transformation
$f = \bar f + Vz$ yields $ \bar f + V B(z^\star, 1)
    \;\subseteq\;
    S'
    \;\subseteq\;
    \bar f + V B(z^\star, 4\nu + 1)$. 
Since $V$ acts only on the displacement variable, we also have the equivalent
representation $
    \bar f + V B(z^\star, r)
    = 
    f^{\star} + V B(0, r)$.
This ellipsoidal sandwiched approximation provides a computationally tractable
geometric surrogate for the true feasible set $S'$. Figure~\ref{fig:geometry} geometrically illustrates how the OD flow
uncertainty set emerges from the nullspace of the link–OD incidence matrix and
how the ellipsoidal approximation captures the feasible variation directions.

The representation $f = \bar f + Vz$ highlights that all admissible OD flows in
$S'$ differ only through movements in the nullspace directions encoded by $z$.
These directions do not affect the physical constraint $L = Xf$, since $XV = 0$,
and, when $z$ remains within the ball $B(0,r)$, the resulting $f$ automatically
respects the box constraints $0 \le f \le U$.  
Thus, $z$ captures the “degrees of freedom’’ available for adjusting $f$ without
violating either the physical conservation constraints or the OD bounds.
This flexibility allows downstream decision models to choose $z$ in a way that
optimizes performance while remaining within the feasible ellipsoidal region.
  
In particular, the ellipsoidal constraint
\begin{equation}
    z^\top Hz\le r^2,
\end{equation}
where $r\ge1$ controls the radius of the search region, provides a compact and tractable characterization of the controllable degrees of freedom available for downstream optimization. The choice $r=1$ guarantees that the ellipsoid is contained in $S'$. As the nullspace dimension grows, however, the unit ellipsoid may capture only a small portion of the admissible OD variation and thus become overly conservative. We therefore treat $r$ as a tuning parameter: increasing $r$ expands the ellipsoidal search region and allows a broader range of decision-relevant demand patterns to be explored. Although $r>1$ relinquishes the automatic containment guarantee, the nullspace construction continues to preserve flow conservation because $XV=0$; only the lower and upper bounds on OD demand may be violated. Such violations can be tolerated when modest, while the original bounds can be imposed explicitly whenever strict feasibility is required. Appendix~\ref{app:ellipsoid-scale} analyzes the effect of nullspace dimension and discusses the calibration of $r$.

This controlled expansion is useful for decision aware estimation because the goal is to reduce downstream regret rather than preserve statistical fidelity alone. The resulting estimate therefore need not be unbiased, and small violations of the data-derived OD bounds may be justified if they improve downstream decisions. For example, suppose an intervention is triggered when estimated demand exceeds 25. If failing to intervene when actual demand exceeds this threshold is more costly than intervening unnecessarily, a modest upward bias near the threshold may be preferable to a comparable downward bias. A larger $r$ provides the flexibility for such decision driven adjustments, while the regret objective determines their direction and magnitude.


\subsection{Optimization Formulation for Decision Aware OD Estimation} \label{Framework}



Consider the following decision problem that takes OD flow $f$ as an input:
\begin{equation}
    \min_{w \in \mathcal{W}} f^\top w,
    \label{eq:lp_decision}
\end{equation}
where $w$ denotes the downstream decision and $\mathcal{W}$ is a nonempty compact polyhedron defined by the linear operational constraints of the downstream problem. We denote the optimal objective value of this problem by
$c(f)=\min_{w \in \mathcal{W}} f^\top w$, and denote by
$w(f) \in \arg\min_{w \in \mathcal{W}} f^\top w$ an optimal decision under $f$. Note that, by duality, $f$ can equivalently appear either as coefficients in the objective function or as right-hand-side constants in the constraints. This formulation captures many transportation applications in which OD flows drive payoffs, costs, or service levels, including assignment problems (e.g., allocating vehicles to OD pairs), matching problems (e.g., pairing demand and supply across zones), and transportation or freight flow problems with fixed capacity constraints.

If the true OD flow is $f_{\rm true}$ but the downstream optimization problem is solved using an estimated OD flow vector $f$, the resulting decision generally yields a suboptimal objective value. The associated loss is measured by the regret defined in~\eqref{eq-regret}, which captures the degradation in decision quality induced by OD estimation error. In practice, $f_{\rm true}$ is fixed but unobserved. Suppose that prior information provides an estimate $f_0$ of the true OD flow. We model the unknown discrepancy between $f_0$ and $f_{\rm true}$ by a random error $\varepsilon$ with distribution $\mathcal{P}$, and represent the uncertain true OD flow as
$$
F = f_0 + \varepsilon,
\qquad
\varepsilon \sim \mathcal{P}.
$$
Here, $f_0$ is fixed and observed, whereas $\varepsilon$ and hence $F$ are random. For now, we assume that the error distribution $\mathcal{P}$ is known. Appendix~\ref{app:local-neighborhood} describes how $\mathcal{P}$ can be constructed from historical prediction residuals and contextual information, including historical demand and socioeconomic features.


Decision aWare Estimation (DWE) therefore seeks to minimize the expected regret under the uncertainty in the true OD flow:

\begin{equation}
\tag{DWE-original}
\min_{g(z) \in S'}
\; \mathbb{E}_{\varepsilon \sim \mathcal{P}}
    \left[
        R\bigl(f_0+\varepsilon,\; g(z)\bigr)
    \right].
\label{eq:E_regret_error_original}
\end{equation}
Recall that $S'$ is the feasible OD set induced by link-flow conservation and OD bounds, and that $g(z)=\bar f+Vz$ parameterizes a candidate OD estimate in $S'$.

Within the ellipsoidal approximation of the feasible region, the DWE formulation becomes
\begin{equation}
\tag{DWE-Ellipsoid}
\min_{z : z^\top H z \le r^2}
\; \mathbb{E}_{\varepsilon \sim \mathcal{P}}
    \left[
        R\bigl(f_0+\varepsilon,\; g(z)\bigr)
    \right].
\label{eq:E_regret_error}
\end{equation}

For applications where tail risk plays an important role, a more conservative formulation replaces the expectation with the superquantile, also known as Conditional Value at Risk (CVaR), where $\beta\in(0,1)$ denotes the confidence level:
\begin{equation}
\min_{z : z^\top H z \le r^2}
\; \mathrm{CVaR}_{\beta,\, \varepsilon \sim \mathcal{P}}
    \left[
        R\bigl(f_0+\varepsilon,\; g(z)\bigr)
    \right].
\label{eq:CVaR_regret_error}
\end{equation}

\subsection{Surrogate Regret Formulations}\label{SOL-SPO}

The regret defined in~\eqref{eq-regret} is generally nonconvex in the estimated OD flow $f$. Therefore, directly minimizing its expectation or a risk measure with respect to $f$ (or equivalently with respect to $z$ through $f=\bar f+Vz$) leads to a nonconvex optimization problem.

\cite{elmachtoub2022smart} shows that, for linear optimization problems, the regret admits a tractable convex upper bound known as the \emph{SPO\(+\) loss}. We follow their construction and retain the auxiliary scaling parameter $\alpha>0$ that arises in the derivation of the SPO$+$ upper bound, yielding a parameterized surrogate denoted by $\ell_\alpha$. We discuss the role and choice of $\alpha$ later in this section. Let $f^\ast$ denote a realization of the uncertain true OD flow $F=f_0+\varepsilon$. Given $f^\ast$ and a candidate OD estimate $f$, the parameterized surrogate is defined as
\begin{equation}
    \ell_\alpha(f^\ast,f)
    :=
    \max_{w\in\mathcal{W}}
        (f^\ast-\alpha f)^\top w
    + \alpha f^\top w(f^\ast)
    - {f^\ast}^\top w(f^\ast).
    \label{eq:spo_plus_ours}
\end{equation}

The following proposition summarizes the properties of $\ell_\alpha$ in our notation. Appendix~\ref{APP:PP2} provides the proof.

\begin{proposition}[Properties of $\ell_{\alpha}$]
\label{prop:spo_plus_ours}
For any $\alpha>0$ and any realized OD flow vector $f^\ast$, the surrogate $\ell_{\alpha}$ satisfies:
\begin{enumerate}
    \item \textbf{Upper bound on regret:}
    $$
        R(f^\ast,f)
        \;\le\;
        \ell_{\alpha}(f^\ast,f)
        \qquad\text{for all } f\in\mathbb{R}^n;
    $$
    \item \textbf{Convexity:}
    $\ell_{\alpha}(f^\ast,f)$ is a convex function of $f$;
    \item \textbf{Subgradient:}
    for any $f$,
    $$
        \alpha \bigl(w(f^\ast) - w(\alpha f - f^\ast)\bigr)
        \;\in\;
        \partial_f\,
        \ell_{\alpha}(f^\ast, f).
    $$
\end{enumerate}
\end{proposition}

Proposition~\ref{prop:spo_plus_ours} establishes several fundamental properties of the $\ell_{\alpha}$ surrogate, including its tractability and its ability to provide an upper bound on the true regret associated with a downstream linear decision problem. 

The classical SPO$+$ construction of \cite{elmachtoub2022smart} sets $\alpha=2$ because, within the empirical risk minimization (ERM) framework, this choice yields a convex surrogate that is Fisher consistent, meaning that, under the stated conditions, minimizing the population SPO$+$ risk also minimizes the true SPO risk. However, our approach does not rely on such distributional consistency. Rather than learning a direct prediction $f$ through ERM, we optimize the parameterized surrogate over a physically constrained feasible OD space. Under this paradigm, the desirable value of $\alpha$ is not dictated by Fisher consistency of an ERM estimator, but by the quality of the resulting surrogate and its downstream decision performance, thereby allowing greater flexibility in its choice.

The surrogate objective $\ell_{\alpha}(f^\ast,f)$ is affected by two distinct
sources of discrepancy: (i)~\emph{surrogate approximation}, arising from
replacing the original regret with a tractable convex upper bound; and
(ii)~\emph{uncertainty perturbation}, arising because the surrogate is evaluated
at a realization $f^\ast$ of the uncertain OD flow rather than at the
unobserved true flow $f_{\rm true}$. These two effects interact with the choice
of $\alpha$. The following proposition provides an explicit bound that
characterizes this tradeoff.

\begin{proposition}[Tradeoff Bound for the $\ell_{\alpha}$ Surrogate]
\label{prop:alpha-tradeoff}
Let $f^\ast$ be a realization of the uncertain OD flow $F$. Suppose that $\mathcal W$ is compact and that there exist constants $U>0$ and $B_w>0$ such that $\|f\|,\|f^\ast\|,\|f_{\rm true}\|\le U$ and $\|w\|\le B_w$ for all $w\in\mathcal W$. Further suppose that, for the candidate OD estimate $f$, there exists $\kappa>0$ such that $f^\top(w-w(f))\ge\kappa\|w-w(f)\|$ for all $w\in\mathcal W$.

For any $\alpha>0$, the discrepancy between the surrogate and the true regret
admits the decomposition
$$
\ell_\alpha(f^\ast,f)-R(f_{\rm true},f)=
\underbrace{\ell_\alpha(f_{\rm true},f)-R(f_{\rm true},f)}
_{\text{surrogate approximation}}+
\underbrace{\ell_\alpha(f^\ast,f)-\ell_\alpha(f_{\rm true},f)}
_{\text{uncertainty perturbation}}.
$$
Moreover,
\begin{equation}
\ell_\alpha(f^\ast,f)-R(f_{\rm true},f)
\le
\frac{2U^2B_w}{\kappa\alpha}
+
2\alpha B_w
\left(
\|f-f_{\rm true}\|+\|f^\ast-f\|
\right)
+
2B_w\|f^\ast-f_{\rm true}\|.
\label{eq:alpha-tradeoff-bound}
\end{equation}
If $\|f-f_{\rm true}\|+\|f^\ast-f\|>0$, the $\alpha$-dependent part of
the bound is minimized at
\begin{equation}
\alpha^\ast
=
\frac{U}{
\sqrt{
\kappa\left(
\|f-f_{\rm true}\|+\|f^\ast-f\|
\right)
}}.
\label{eq:oracle-alpha}
\end{equation}
\end{proposition}

The detailed proof is given in Appendix \ref{app:Pro2}.

\paragraph{Discussion.}
Proposition~\ref{prop:alpha-tradeoff} shows that $\alpha$ balances surrogate
approximation against sensitivity to OD flow discrepancies. In \eqref{eq:alpha-tradeoff-bound}, the approximation term decreases as $1/\alpha$, whereas the discrepancy-dependent term increases linearly with $\alpha$. Hence, a small $\alpha$ yields a looser regret approximation, while a large $\alpha$ makes the surrogate more sensitive to differences among the candidate, realized, and true OD flows. The remaining uncertainty term is
independent of $\alpha$. Equation~\eqref{eq:oracle-alpha} summarizes this balance. Larger OD-flow
discrepancies favor a smaller $\alpha$ because their contribution to the bound
is amplified linearly by $\alpha$. In contrast, a smaller $\kappa$ favors a
larger $\alpha$: when the downstream optimum is less sharply separated, a
larger scaling factor is needed to control the surrogate-approximation term. Since $f_{\rm true}$ is unobserved, $\alpha^\ast$ serves as an oracle benchmark; in practice, $\alpha$ is selected based on downstream decision
performance in Section~\ref{DAODE-EXP}.

By replacing $R(f_0+\varepsilon,g(z))$ in the objectives of \eqref{eq:E_regret_error_original} and \eqref{eq:E_regret_error} with its convex surrogate $\ell_{\alpha}(f_0+\varepsilon,g(z))$, and noting
that $g(z)=\bar f+Vz$ is affine in $z$, the resulting optimization problems
are convex in $z$ and can be solved using standard convex optimization methods.

\begin{theorem}[Convexity and Subgradient of the Expectation Formulation]
\label{T1}
Let $H\in\mathbb{R}^{d\times d}$ be positive definite, and consider
$$
\min_{z\in\mathbb{R}^d:\;z^\top H z\le r^2}
\phi(z):=\mathbb{E}_{\varepsilon\sim\mathcal P}\left[\ell_{\alpha}(f_0+\varepsilon,g(z))\right].
$$
Assume that $\ell_{\alpha}(f_0+\varepsilon,g(z))$ is integrable for every feasible $z$, and that the subgradient expression below is integrable with respect to $\varepsilon\sim\mathcal P$. Then:
\begin{enumerate}
    \item $\phi(z)$ is convex in $z$, and the optimization problem is a convex optimization problem;
    \item a subgradient of $\phi$ at $z$ is given by
    $$
    \mathbb{E}_{\varepsilon\sim\mathcal P}
    \left[
    \alpha V^\top
    \left(
    w(f_0+\varepsilon)
    -
    w\bigl(\alpha g(z)-f_0-\varepsilon\bigr)
    \right)
    \right]
    \in \partial\phi(z).
    $$
\end{enumerate}
\end{theorem}

Theorem~\ref{T1} shows that the expectation formulation is a convex optimization problem. Since $\ell_{\alpha}(f^\ast,f)$ is convex but generally nonsmooth in its second argument, $\phi(z)$ is convex and may also be nondifferentiable. By Proposition~\ref{prop:spo_plus_ours}, for each realization $f_0+\varepsilon$, a subgradient with respect to the second argument is $\alpha[w(f_0+\varepsilon)-w(\alpha g(z)-f_0-\varepsilon)]$. Since $g(z)$ is affine, multiplication by $V^\top$ gives the corresponding
subgradient with respect to $z$, and taking expectations yields the expression
in Theorem~\ref{T1}. An analogous argument applies to CVaR; see
Appendix~\ref{app:cvar_extension} for a formal statement and proof.

\subsection{Equivalent Reformulations and Computational Solution}
\label{sec:computational_formulation}

We next show that the surrogate DWE formulations admit standard deterministic
optimization reformulations and can therefore be solved directly using
off-the-shelf optimization software. To obtain a finite-dimensional computational representation, we approximate the
prior distribution of the true OD flow by a discrete scenario distribution.
Specifically, let
$\tau_k:=f_0+\varepsilon_k$, $k=1,\ldots,K$, denote the representative
OD-flow realizations, with associated probabilities $p_k$, where
$\sum_{k=1}^K p_k=1$. This yields the discrete approximation $\widehat{\mathcal P}=\sum_{k=1}^K p_k\,\delta_{\tau_k}$, where $\delta_{\tau_k}$ denotes a point mass at $\tau_k$. We further define $\bar w
:=
\mathbb{E}_{\tau\sim\widehat{\mathcal P}}[w(\tau)]
=
\sum_{k=1}^K p_k w(\tau_k),$, which is the probability-weighted average of the optimal downstream decisions
induced by the prior OD-flow realizations.

Under the finite-scenario representation of the prior distribution, substituting
the definition of $\ell_\alpha(\tau_k,f)$ in \eqref{eq:spo_plus_ours} yields
\begin{equation}
\begin{aligned}
\sum_{k=1}^K p_k\,\ell_\alpha(\tau_k,f)
=
&\sum_{k=1}^K p_k
\max_{w\in\mathcal W}
(\tau_k-\alpha f)^\top w
+\alpha f^\top \bar w -
\sum_{k=1}^K p_k\,\tau_k^\top w(\tau_k).
\end{aligned}
\label{eq:scenario_surrogate}
\end{equation}
The last term is independent of the estimation decision $f$ and can therefore
be omitted without affecting the optimizer.

We now exploit the linear structure of the downstream decision problem.
Suppose that its feasible region admits the general polyhedral representation
$$
\mathcal W
=
\left\{
w\in\mathbb R^n:
Aw\le b,\;
Bw=d
\right\},
$$
where the equality constraints capture, for example, flow-conservation or
balance requirements, while capacity constraints, resource restrictions, and
variable bounds can be incorporated into $Aw\le b$.

For each prior scenario $k$, the inner optimization problem is therefore
$$
\max_{w}
\left\{
(\tau_k-\alpha f)^\top w:
Aw\le b,\;
Bw=d
\right\}.
$$
Let $\lambda_k\ge0$ denote the dual variables associated with the inequality
constraints and let $\mu_k$ denote the unrestricted dual variables associated
with the equality constraints. Assuming the downstream problem is feasible
and bounded, strong linear programming duality gives
$$
\max_{w}
\left\{
(\tau_k-\alpha f)^\top w:
Aw\le b,\;
Bw=d
\right\}
=
\min_{\lambda_k\ge0,\;\mu_k}
\left\{
b^\top\lambda_k+d^\top\mu_k:
A^\top\lambda_k+B^\top\mu_k
=
\tau_k-\alpha f
\right\}.
$$
Thus, the nested optimization in the surrogate loss can be eliminated,
yielding a single-level deterministic equivalent formulation.

For the original feasible OD region $S'$, DWE-original is equivalent, up to an additive constant independent of the
decision variables, to
\begin{equation}
\begin{aligned}
\min_{f,\{\lambda_k,\mu_k\}_{k=1}^K}\quad
    & \alpha f^\top \bar w
      +
      \sum_{k=1}^K p_k
      \left(
      b^\top\lambda_k+d^\top\mu_k
      \right) \\
\text{s.t.}\quad
    & A^\top\lambda_k+B^\top\mu_k
      =
      \tau_k-\alpha f,
      && k=1,\ldots,K,\\
    & \lambda_k\ge0,
      && k=1,\ldots,K,\\
    & Xf=L,\\
    & 0\le f\le U.
\end{aligned}
\tag{DWE-original-DE}
\label{eq:lp_dwe}
\end{equation}
The objective and all constraints in \ref{eq:lp_dwe} are linear.
Hence, the finite-scenario surrogate formulation of DWE-original is a
linear program.

The same dual reformulation applies to DWE-Ellipsoid. Substituting $f=g(z)=\bar f+Vz$ yields
\begin{equation}
\begin{aligned}
\min_{z,\{\lambda_k,\mu_k\}_{k=1}^K}\quad
    & \alpha(\bar f+Vz)^\top\bar w
      +
      \sum_{k=1}^K p_k
      \left(
      b^\top\lambda_k+d^\top\mu_k
      \right) \\
\text{s.t.}\quad
    & A^\top\lambda_k+B^\top\mu_k
      =
      \tau_k-\alpha(\bar f+Vz),
      && k=1,\ldots,K,\\
    & \lambda_k\ge0,
      && k=1,\ldots,K,\\
    & z^\top H z\le r^2.
\end{aligned}
\tag{DWE-Ellipsoid-DE}
\label{eq:qcqp_dwe}
\end{equation}
Since $H\succ0$, the ellipsoidal constraint is convex, while the objective
and all remaining constraints are linear. Thus, \ref{eq:qcqp_dwe} can be viewed as a special case of a convex quadratically constrained quadratic program (QCQP), with a linear objective function and a single convex quadratic constraint. Moreover, letting $H=R^\top R$ denote a Cholesky factorization, the
ellipsoidal constraint can equivalently be expressed as $\|Rz\|_2\le r$. Hence, \ref{eq:qcqp_dwe} also admits an equivalent second-order cone programming
(SOCP) representation.

The resulting formulations belong to standard classes of convex optimization problems. \ref{eq:lp_dwe} reduces to an LP, whereas \ref{eq:qcqp_dwe} can be
formulated as either a convex QCQP or an SOCP. Both formulations can be
handled directly by modern optimization solvers, thereby avoiding the need
for a customized nonsmooth optimization algorithm.

\section{Numerical Experiments}\label{DAODE-EXP}

Numerical experiments based on three test networks of varying sizes are conducted to test and validate the proposed decision aware OD estimation framework.  Besides checking the accuracy and scalability of our method, we also examine how decision regret varies with model misspecification and data noise, and show that the proposed framework effectively reduces decision error and adapts the surrogate parameter $\alpha$ as the error structure changes. 


\subsection{An Illustrative Toy Example for Motivating Decision-awareness}\label{subsec:illustrative_3d_example}

This simple numerical example is designed to motivate the need for decision awareness and to illustrate the interaction between various pulling sources, including prior information presented by historical samples, physical feasibility based on current sensor observations, and downstream decision quality, in an underdetermined OD estimation problem.

\begin{figure}[t]
    \centering
    \includegraphics[width=0.75\linewidth]{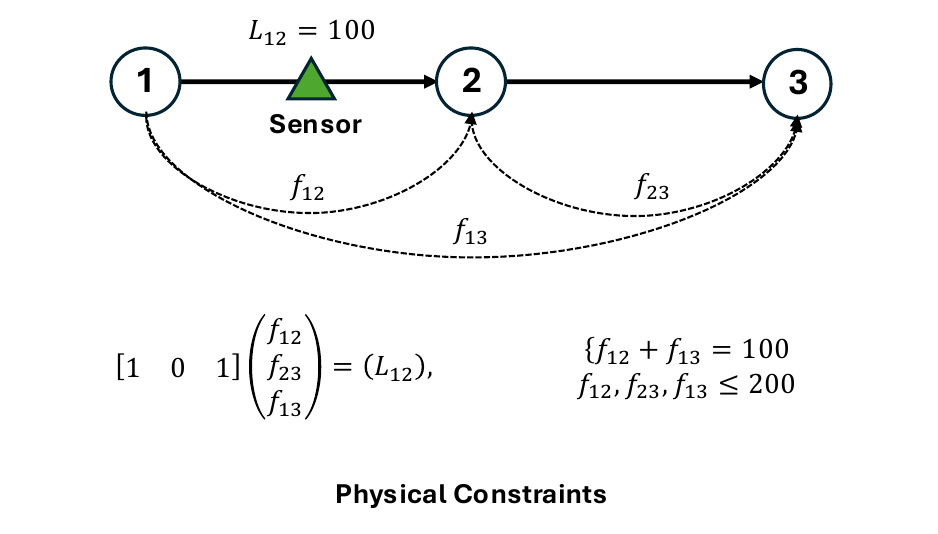}
    \caption{
    Illustrative three-node network with three OD flows, a single sensor observation, and the induced physical constraint.
    }
    \label{fig:3Dexample}
\end{figure}

Consider a toy network with three nodes, three OD flows, and two directed links, as shown in Figure~\ref{fig:3Dexample}. Let $f=(f_{12},f_{23},f_{13})^\top$ denote the OD demand vector. Suppose that only link $(1,2)$ is instrumented with a sensor, and the observed link flow is $L_{12}=100.$ Under the path structure in Figure~\ref{fig:3Dexample}, this observation implies the linear constraint $f_{12}+f_{13}=100.$ In addition, assume simple capacity bounds $0\le f_{12},f_{23},f_{13}\le 200.$ These constraints form the feasible OD set, and any point within this feasible set aligns with the observed data.  

Now let us illustrate how OD estimation affects downstream decisions.  Consider the following simple decision problem:
\begin{equation}
\begin{aligned}\label{eq:toy_decision_problem}
\min_{\mathbf{w}\in\mathbb{R}^3}\quad & f_{12}w_1+f_{23}w_2+f_{13}w_3 \\
\text{s.t.}\quad 
& w_1+w_2+w_3=1,\\
& w_1-w_3=0,\\
& 0\le w_1,w_2,w_3\le 1.
\end{aligned}
\end{equation}

With symmetry between $w_1$ and $w_3$, problem~\eqref{eq:toy_decision_problem} reduces to comparing two effective actions. The derived decision rule is
$$
\begin{cases}
w=(0,1,0), & \text{if } f_{23}\le \dfrac{f_{12}+f_{13}}{2},\\[6pt]
w=\left(\dfrac{1}{2},0,\dfrac{1}{2}\right), & \text{if } f_{23}>\dfrac{f_{12}+f_{13}}{2}.
\end{cases}
$$
For OD estimates that satisfy the sensor information $f_{12}+f_{13}=100$, this decision boundary reduces to $f_{23}=50$. Therefore, the downstream action depends only on whether $f_{23}$ lies below or above the threshold value $50$. This is precisely the type of setting in which a decision aware estimator can differ sharply from a prediction-oriented one: not all directions of OD error are equally consequential, and errors that move the estimate across the decision boundary are much more important than errors that remain decision-neutral.

We set the true OD vector to $f^{\star}=(35,49,65)^\top$, which satisfies the physical constraint $f_{12}+f_{13}=100$. Under the true OD vector, $\min\{f_{23}^{\star},50\}=\min\{49,50\}=49$, so the optimal solution is the $f_{23}$-driven action, i.e., $w=(0,1,0)$. We then construct a dataset of $300$ historical OD samples and consider the sample mean as the simplest benchmark predictor. To make the distinction between prediction error and decision impact transparent, we construct two subgroups of historical samples with comparable Euclidean deviation from the true OD vector but sharply different decision impacts: 
\begin{itemize}
    \item The first subgroup contains $200$ \emph{safe} historical samples generated around the center $f^{\star}+(10,0,-10)^\top$. Specifically, for each sample we add an integer-valued perturbation $(\varepsilon_1,\varepsilon_2,\varepsilon_3)$, where $\varepsilon_1$ and $\varepsilon_3$ are constructed so that $\varepsilon_3=-\varepsilon_1$, and $\varepsilon_2$ is a relatively small perturbation around zero. These samples are intentionally displaced in a direction that mainly redistributes mass between $f_{12}$ and $f_{13}$ while leaving the decision-relevant component $f_{23}$ nearly unchanged. As a result, although their mean squared deviation from the truth is non-negligible, their effect on the downstream decision is limited.
    \item The second subgroup contains $100$ \emph{risky} historical samples generated around the center $f^{\star}+(0,10\sqrt{2},0)^\top$, with moderate perturbations around that point. This group is constructed so that its Euclidean distance from the truth is of a similar magnitude to that of the safe group, but the displacement occurs primarily in the sensitive direction associated with $f_{23}$. Consequently, these samples are much more likely to move the estimate across the decision threshold and induce the wrong action.
\end{itemize}

This construction is deliberate. The safe and risky samples have nearly identical average Mean Squared Error (MSE) relative to the true OD vector---$67.27$ for the safe group and $65.35$ for the risky group---and are therefore difficult to distinguish under a standard prediction metric. Yet their decision implications are very different. Most samples remain consistent with the correct downstream action, whereas a smaller subset is systematically shifted toward the wrong side of the decision boundary and thus behaves like decision-distorting outliers.  These samples do not stand out in terms of MSE and do not dominate the dataset numerically, yet they are sufficient to shift a conventional predictor across the decision threshold and induce the wrong decision.

\begin{figure}[t]
    \centering
    \includegraphics[width=\linewidth]{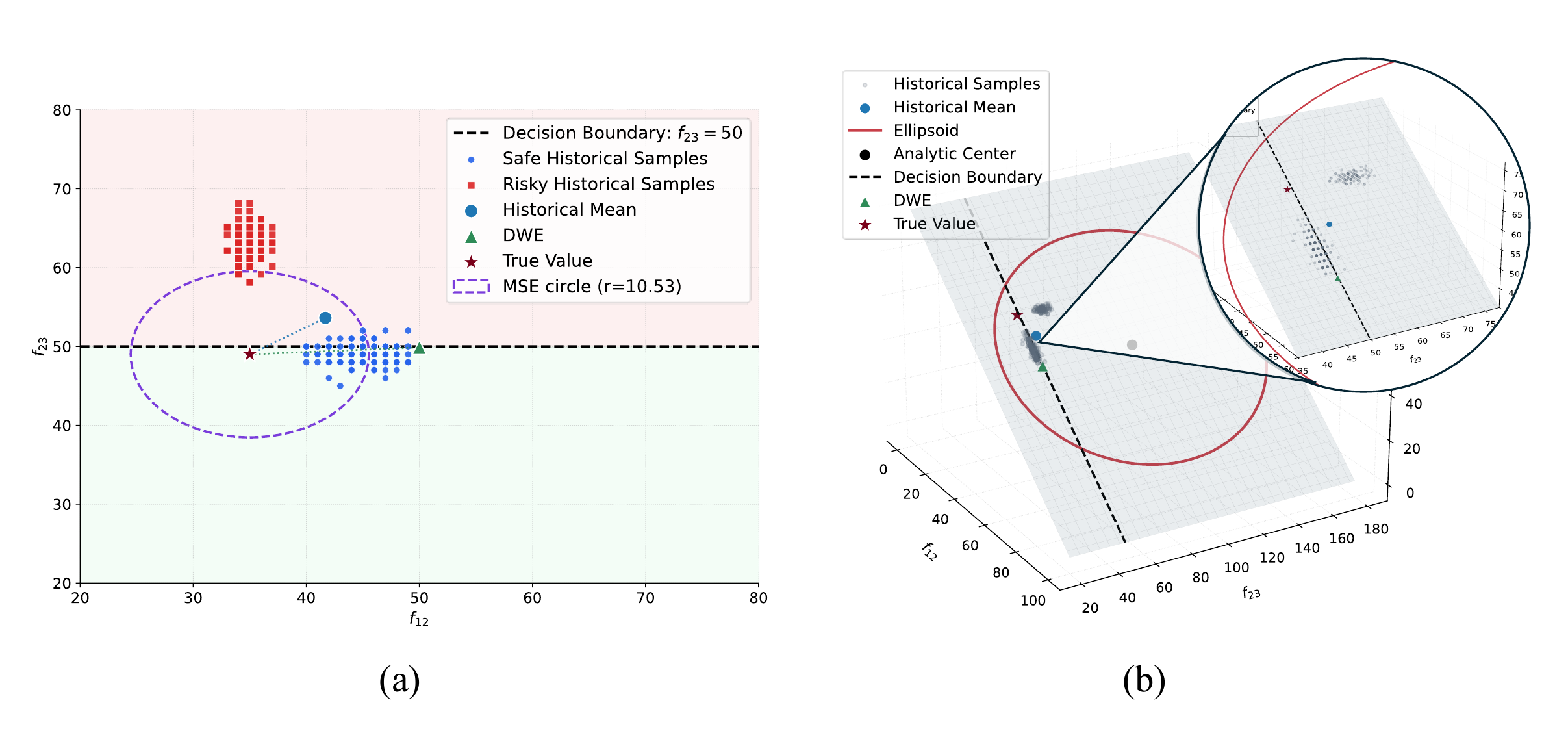}
    \caption{Comparison of the mean predictor and the DWE solution under decision-distorting samples: (a) reduced two-dimensional view on the feasible manifold; (b) corresponding three-dimensional geometric illustration.}
    \label{fig:3DexampleEX1}
\end{figure}

This distinction is reflected in both the aggregate prediction results and the geometry shown in Figure~\ref{fig:3DexampleEX1}. Using the sample mean of the entire group of historical samples as a benchmark predictor, yields $\widehat{f}^{\mathrm{mean}}=(41.69,\;53.62,\;58.30)^\top$. By contrast, the proposed Decision Aware Estimation (DWE) method returns $\widehat{f}^{\mathrm{DWE}}=(50.00,\;49.90,\;50.00)^\top$. In prediction terms, the sample mean is substantially closer to the ground truth: its MSE is $36.96$, whereas the DWE estimate has a much larger MSE of $150.27$. However, the performance reverses once decision quality is taken into account. Because $\widehat f_{23}^{\mathrm{mean}}>50$, the sample mean induces the wrong decision; in contrast, because $\widehat f_{23}^{\mathrm{DWE}}<50$, the DWE estimate recovers the correct decision.

Figure~\ref{fig:3DexampleEX1} illustrates the example from two complementary perspectives. In the reduced 2-D view in panel (a), the downstream decision depends only on whether $f_{23}$ lies above or below the threshold of $50$. Although most samples lie in the decision-consistent region, a small cluster of decision-distorting samples pulls the historical mean (blue point) across the boundary, despite its smaller MSE relative to the true OD vector (black star). The DWE solution (green triangle), in contrast, remains in the decision-consistent region. Panel (b) reveals the underlying geometry: the historical mean is pulled toward the center of the data cloud on the sensor-induced feasible manifold, improving data fidelity but crossing the projected decision boundary, whereas DWE optimizes the decision aware objective within the geometrically approximated feasible region and shifts the estimate toward a decision-robust region. Thus, lower prediction error does not necessarily translate into better operational performance.

Next, we examine a situation where historical samples violate the physical feasibility induced by hard data. We generate 300 historical OD samples:
\begin{itemize}
    \item $200$ \emph{regular} samples satisfying the physical relation $f_{12}+f_{13}=100$, with $f_{12}\in[25,45], f_{23}\in[45,50], f_{13}=100-f_{12}.$ 
    \item $100$ \emph{infeasible} samples for which $f_{12}$ and $f_{23}$ remain in the same range, but $f_{13}$ is shifted downward so that $f_{13}=(100-f_{12})-\xi, \xi\in[20,30].$ As a result, these contaminated historical samples no longer lie on the physically feasible manifold. 
\end{itemize}
This construction captures an important failure mode in OD learning: in practice, historical data may contain observations that are inconsistent with the current physical state of the network. 


\begin{figure}[t]
    \centering
    \includegraphics[width=\linewidth]{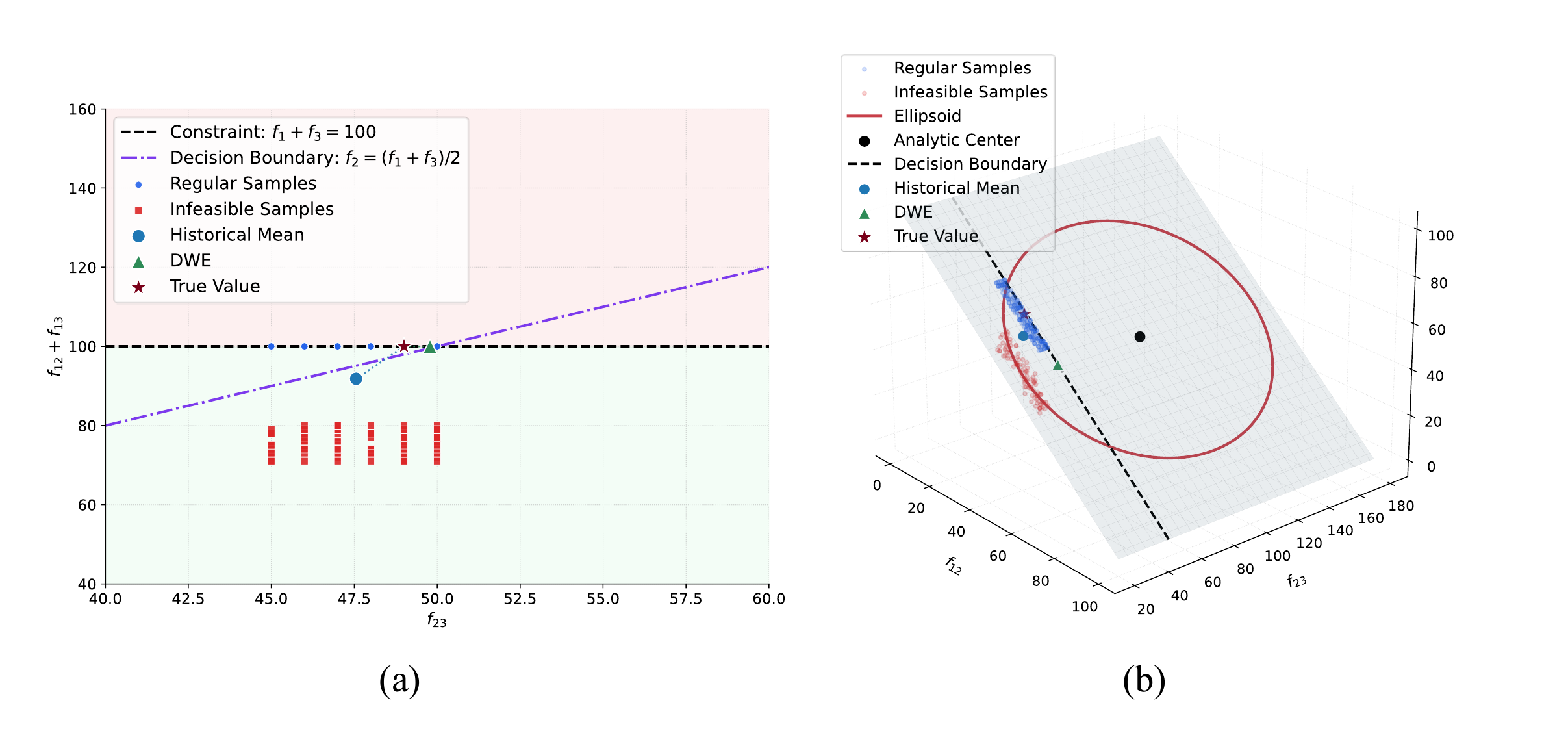}
    \caption{Comparison of the mean predictor and the DWE solution under infeasible samples: (a) reduced two-dimensional view showing physical feasibility and the decision boundary; (b) corresponding three-dimensional geometric illustration.}
    \label{fig:3DexampleEX2}
\end{figure}

Figure~\ref{fig:3DexampleEX2} illustrates the effect of physically infeasible samples from two complementary perspectives. In the reduced 2-D view in panel (a), $f_{12}+f_{13}=100$ defines the sensor-implied feasibility condition, while $f_{23}=(f_{12}+f_{13})/2$ defines the decision boundary. The infeasible samples (red points) pull the historical mean away from the physically feasible region, whereas DWE optimizes the decision aware objective directly over the feasible manifold, yielding a solution close to the true value while satisfying the physical constraints by construction. Panel (b) reveals the corresponding 3-D geometry: infeasible samples lie systematically off the manifold and distort the raw data cloud, making the historical mean an unreliable summary, while DWE remains anchored to the physically feasible region. Thus, physical constraints protect the estimator from sample errors that would otherwise pull it away from the structure of the current system.

This example illustrates the limitation of purely data-driven estimation. Historical observations are generated under heterogeneous network states, demand conditions, and behavioral responses, and thus populate a broad, high-dimensional space. However, once current observations and network conditions are imposed, the true OD vector is confined to a smaller feasible region lying on a smaller admissible manifold. A predictor may therefore fit the historical data well in a global sense and yet remain misaligned with the structure of the current system. This misalignment is both geometric and operational: some deviations are innocuous, while others are amplified through the downstream optimization, leading to substantial decision regret. The objective is not to reproduce historical patterns, but to extract an estimate that is physically grounded in current conditions and decision-relevant.



\subsection{A Small Network for Benchmarking and Sensitivity Analysis}\label{EX2}

\subsubsection{Network Configuration} 
\begin{figure}[t]
    \centering
    \includegraphics[width=0.75\linewidth]{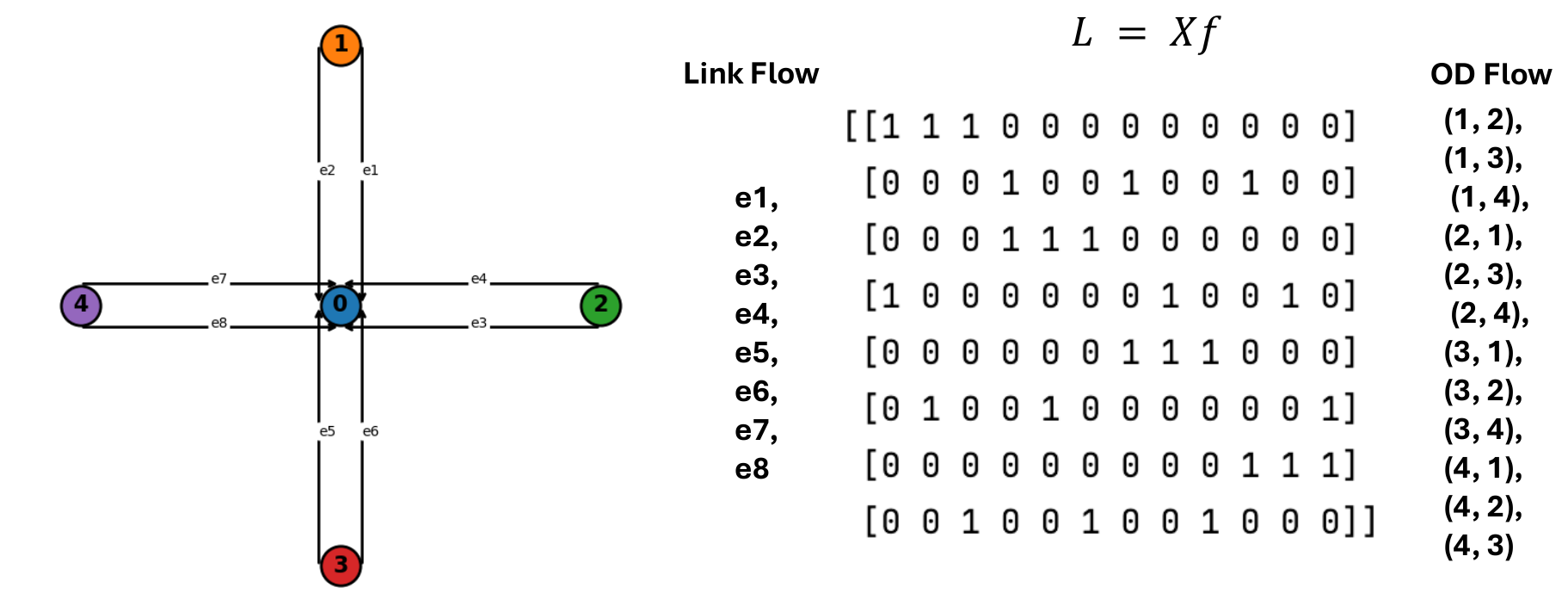}
    \caption{Small network schematic}
    \label{fig:SN}
\end{figure}

Consider a network consisting of four terminal nodes
connected through a single central junction, resulting in eight directed links.
Each terminal node can serve as both an origin and a destination, yielding a
total of 12 OD pairs. The network topology and link
directions are illustrated in Figure~\ref{fig:SN}. Let $f \in \mathbb{R}^{12}_+$ denote the OD flow vector and $L \in \mathbb{R}^8_+$
the corresponding link flow vector. The physical relationship between OD flows
and link flows is encoded by a fixed link-OD incidence matrix $X \in
\{0,1\}^{8 \times 12}$, such that $L = X f$.

\subsubsection{Fabrication of True OD and Historical OD Samples}

We use a polynomial function based on four features to generate historical OD samples and the true OD flows. The case $P=1$ serves as the lowest-complexity benchmark, in which the latent OD index is linear in the features. Larger values of $P$ introduce increasing nonlinear structure in the feature-to-OD mapping. Specifically, for
each polynomial order $P\in\{1,\ldots,6\}$, latent OD flows are generated from
a polynomial transformation of the feature vector and then passed through a
softplus transformation, $Q=\log(1+\exp(Z))$, which guarantees non-negativity while preserving smoothness. Thus, varying $P$ creates OD samples with different levels of complexity and
different degrees of model misspecification for linear prediction models. This
allows us to evaluate how prediction-oriented and decision aware estimators
perform as the historical OD data become increasingly nonlinear and harder to
fit with a linear model.

For each value of \(P\), we generate separate training and testing samples. The
training samples are used to construct prediction-oriented and decision aware
OD estimators, while the testing samples are used to evaluate downstream
decision quality. See Appendix~\ref{app:od_generation} for details of the OD
data-generation process.

\subsubsection{Downstream Decision Problem}

We use the demonstration investment allocation problem described in Appendix~\ref{app:decision_problem} as the downstream decision problem. This problem is formulated as a budget-constrained linear program, in which a transportation authority allocates a fixed investment budget across candidate road segments, and the resulting link improvements reduce OD-level travel times. For a given OD-flow estimate $f$, the downstream problem returns an
optimal investment allocation $x^*(f)$; the corresponding post-investment OD-level travel times are denoted by $w^*(f)$.

To avoid confounding the comparison with random variation from the downstream optimization problem, we evaluate all OD-estimation methods under the same
investment environment within each experimental instance. Specifically, the OD-to-segment incidence matrix, baseline travel times, marginal improvement
costs, and segment-level improvement upper bounds are generated once and then held fixed throughout the method comparison. The investment budget is also fixed and is set to $35\%$ of the maximum feasible investment level. Thus, the downstream LP is identical across methods except for the OD-flow vector used in the objective. This design ensures that differences in downstream performance are driven by the estimated OD vectors rather than by random variation in the underlying investment problem.



\subsubsection{Benchmark Methods}

We organize the benchmark methods according to two dimensions: the source of OD-demand information and the way this information is incorporated into the
downstream decision problem. The first dimension distinguishes different forms of prior OD information available before solving the downstream optimization problem. In particular, we consider three historical-demand priors, denoted by $\mathcal{R}=\{\mathrm{M},\mathrm{LR},\mathrm{Loc}\}$. Here, $\mathrm{M}$ denotes the global historical mean, which uses the average OD vector over the historical sample. $\mathrm{LR}$ denotes a model-based prediction obtained from a linear regression model trained to predict OD flows by minimizing mean squared error; this produces a single point estimate for the current instance. $\mathrm{Loc}$ denotes a local historical-demand prior, which uses multiple historical OD samples associated with covariate values similar to the current instance. In our implementation, this local prior is constructed by averaging the OD vectors of the \(k\)-nearest historical samples in the covariate space.\footnote{More generally, this prior can be viewed as a local regression estimate, and other nonparametric regression methods beyond $k$-nearest neighbors can also be used; see Appendix~\ref{app:local-neighborhood} and \cite{bertsimas2020predictive}.}

We first evaluate these priors under a standard predict-then-optimize (PTO) pipeline. In this class of methods, an OD vector is first estimated without
using the downstream decision objective, and the resulting point estimate is then passed directly to the investment allocation problem. This gives three PTO baselines: $\textsc{PTO-M}, \textsc{PTO-LR}, \textsc{PTO-Loc}$, corresponding respectively to the global historical mean, the linear-regression
prediction, and the local historical-demand prior. These baselines represent the conventional sequential approach in which OD estimation quality and downstream decision quality are optimized separately.

In addition to these demand-side priors, we include an OLS benchmark based on observed link-flow information. This method reconstructs OD flows from observed link flows through a least-squares formulation and then applies the downstream investment model to the reconstructed OD vector. The OLS benchmark therefore represents the information available from the link-flow observation side, in contrast to the historical-demand priors used by the PTO baselines.

We then compare these baselines with two direct-reformulation DWE methods. For each prior-information source $r\in\{\mathrm{M},\mathrm{LR},\mathrm{Loc}\}$, we denote the DWE-original and DWE-Ellipsoid formulations introduced in Section~\ref{sec:computational_formulation} by $\textsc{D-Full-}r$ and $\textsc{D-Geo-}r$, respectively. The former optimizes the SPO+ surrogate over the original link-flow consistent OD feasible region, whereas the latter uses its ellipsoidal approximation. This paired comparison isolates the effect of the feasible-region approximation while holding the prior-information source and decision aware loss fixed.

\subsubsection{Regret Behavior under Increasing Demand Generation Nonlinearity}



This experiment examines how downstream decision regret changes as the polynomial order increases from 1 to 6, thereby introducing progressively stronger nonlinearity into the true demand generation process. This setting allows us to evaluate whether each method remains decision reliable as the actual demand generation mechanism becomes more complex and prediction-centered estimators are increasingly exposed to model misspecification. The comparison is motivated by a central curiosity: Whether and how does model misspecification translate prediction errors into unreliable downstream decisions?  Specifically, we will answer three empirical questions. First, does increasing demand-generation nonlinearity undermine the downstream reliability of decisions produced by prediction-centered methods? Second, conditional on using the same historical demand information, does decision aware estimation provide additional regret reduction beyond standard PTO? Third, are these gains concentrated in average performance, or do they also translate into lower upper-tail regret and greater stability across training samples?

\begin{figure}[t]
    \centering
     \includegraphics[width=\textwidth]{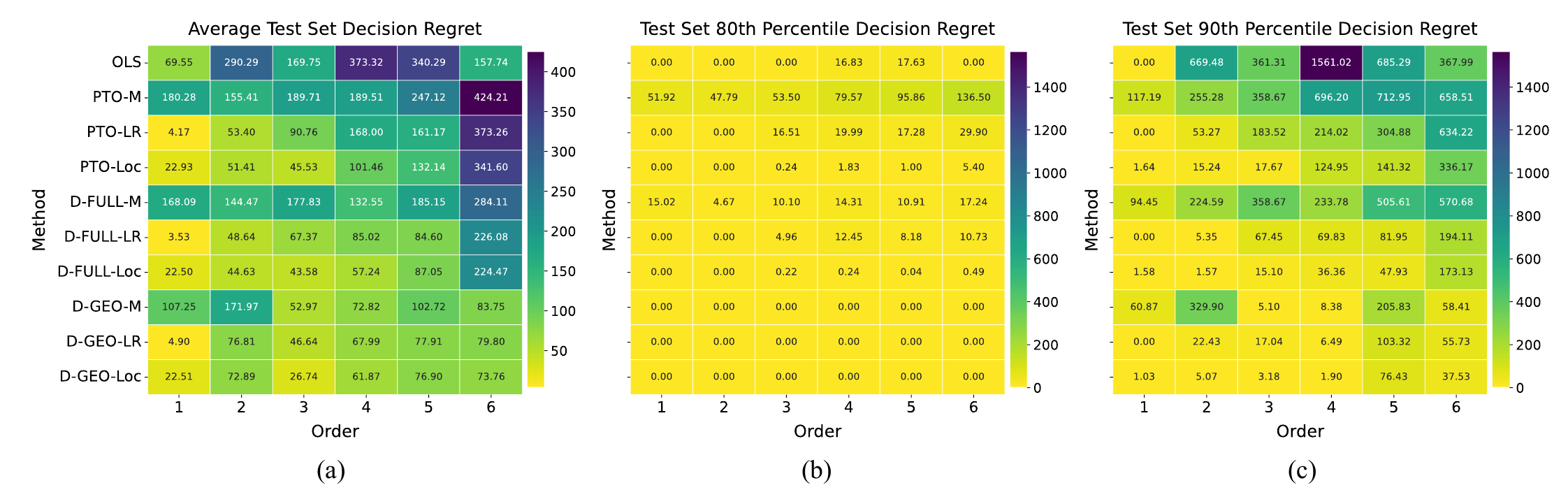}
    \caption{
    Decision regret under increasing nonlinear demand generation.
    Panel (a) reports the average test regret, Panel (b) reports the 80th percentile regret, and Panel (c) reports the 90th percentile regret.
    }
    \label{fig:SN-Order-Heatmap}
\end{figure}

Figure~\ref{fig:SN-Order-Heatmap} summarizes the effect of model misspecification (indicated by increasing demand generation nonlinearity) on downstream decision quality. We first consider
Panel~(a), which reports the average test-set regret. The OLS benchmark provides a useful contrast because it relies only on observed link-flow consistency and does not use any historical OD-demand prior. Its regret varies substantially across polynomial orders. This behavior reflects the ill-posed nature of OD reconstruction merely from aggregate link
flows: many OD vectors can be consistent with the same link-flow observations, but they may induce very different investment decisions, thus leading to unstable downstream regret.

Among the PTO baselines, $\textsc{PTO-M}$ performs poorly because it relies only on a coarse global mean prior and ignores the covariate information of the current instance. Incorporating covariate information substantially improves performance in the low-order regime, reflected by the low regrets produced by $\textsc{PTO-LR}$ and $\textsc{PTO-Loc}$.  For example, $\textsc{PTO-LR}$ produced $4.17$ at order~1, which reflects the advantage of the linear predictor when the true demand-generation process is
close to the assumed model class. However, as the polynomial order increases and model misspecification becomes more severe, both $\textsc{PTO-LR}$ and $\textsc{PTO-Loc}$ deteriorate rapidly.  


We next compare the DWE methods in Panel~(a). The $\textsc{D-Full-}r$ methods provide clear improvements over their PTO counterparts at low and intermediate orders, confirming the value of incorporating downstream decision structure into
the OD-estimation step. However, their performance becomes less stable in the high-order regime. For instance, at order~6, regret remains at $284.11$ for
$\textsc{D-Full-M}$, $226.08$ for $\textsc{D-Full-LR}$, and $224.47$ for $\textsc{D-Full-Loc}$. In contrast, the $\textsc{D-Geo-}r$ methods form the most stable group across orders and prior sources. At order~6, $\textsc{D-Geo-LR}$ and $\textsc{D-Geo-Loc}$ achieve regrets of $79.80$ and $73.76$, respectively. 
The gap between $\textsc{D-Full-}r$ and $\textsc{D-Geo-}r$ is particularly informative. Since both methods use the same decision aware objective, their
difference is driven primarily by the feasible-region representation. The superior stability of $\textsc{D-Geo-}r$ suggests that the geometric
approximation plays a role similar to regularization: it constrains the OD estimate to a more stable and decision-relevant region, thereby preventing the direct decision aware fit over the full feasible set from becoming overly sensitive in high-complexity settings.

Panels~(b) and~(c) further show that the main benefit of the regularized DWE methods is not limited to the average regret. Panel~(b) reports the 80th percentile test-set regret. At this level, differences among the stronger methods become much smaller. In particular, the $\textsc{D-Geo-}r$ methods attain values that are essentially zero across nearly all orders and prior sources, indicating that they incur little or no regret for a large majority of test instances. Thus, much of the average-regret difference in
Panel~(a) is not driven by the bulk of ordinary test cases, but rather by a smaller set of more difficult instances. This interpretation is confirmed by Panel~(c), where the 90th percentile regret reveals much sharper separation across methods. Once attention shifts to the upper tail of the regret distribution, the risk-control advantage of the regularized DWE methods becomes clear. For example, at order~6 under the local prior, the 90th percentile regret is $336.17$ for $\textsc{PTO-Loc}$, $173.13$ for $\textsc{D-Full-Loc}$, and $37.53$ for $\textsc{D-Geo-Loc}$. Under the linear-regression prior at the same order, the corresponding values are $634.22$ for $\textsc{PTO-LR}$, $194.11$ for $\textsc{D-Full-LR}$, and $55.73$ for $\textsc{D-Geo-LR}$. Therefore, the main advantage of the $\textsc{D-Geo-}r$ methods lies not only in reducing average regret, but also in suppressing rare but severe decision failures. This type of tail robustness becomes increasingly important as the true demand-generation mechanism grows more nonlinear.


\begin{figure}[t]
    \centering
    \includegraphics[width=0.95\textwidth]{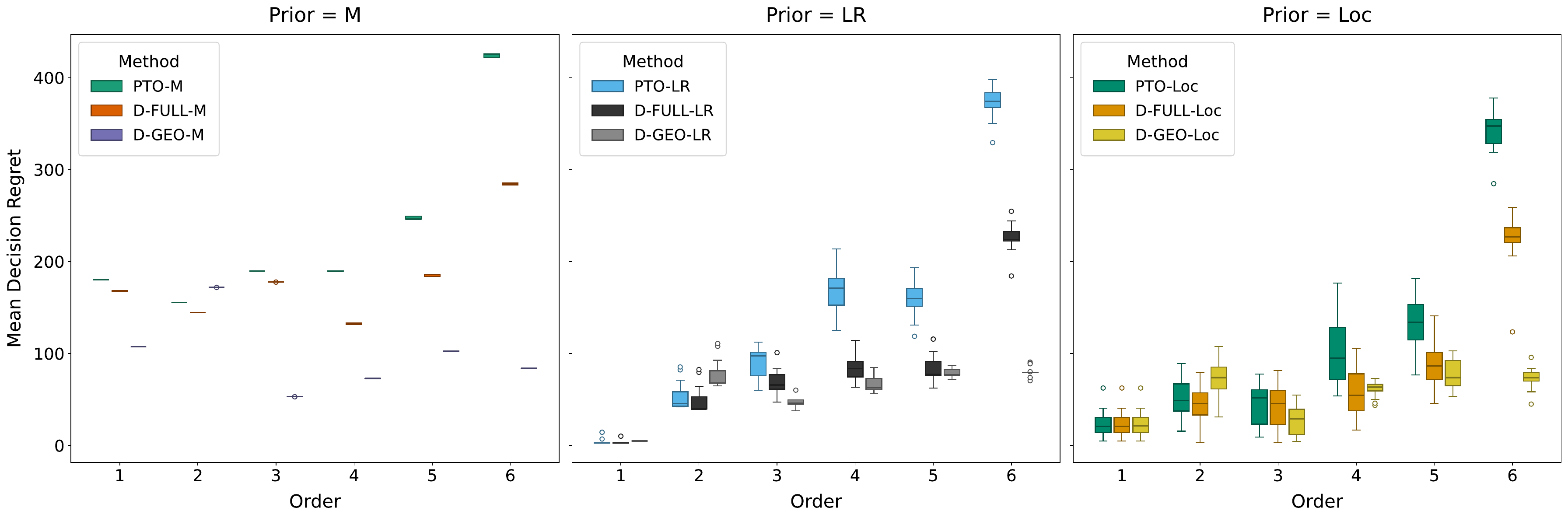}
    \caption{Regret distributions across methods as the polynomial order increases.}
    \label{fig:SN-Order-Boxplot}
\end{figure}

Figure~\ref{fig:SN-Order-Boxplot} complements the heatmap by showing the within-prior distribution of mean test-set regret across training seeds. This figure addresses whether the improvement of the DWE methods is driven by a few favorable training samples or is systematic across training
realizations. In the high-order regimes, $\textsc{PTO-LR}$ and $\textsc{PTO-Loc}$ have not only higher median regret, but also wider boxes,
longer upper whiskers, and more extreme outliers. This indicates that, under stronger nonlinearity, PTO becomes highly sensitive to the particular training sample: some seeds lead to especially poor downstream decisions. In contrast, the $\textsc{D-Geo}$ variants shift the entire seed-level distribution downward and keep it substantially more concentrated. Thus, their improvement is
not driven by a small number of favorable training seeds. Rather, they systematically reduce variability and suppress high-regret training realizations.


\begin{figure}[t]
    \centering
    \includegraphics[width=0.75\textwidth]{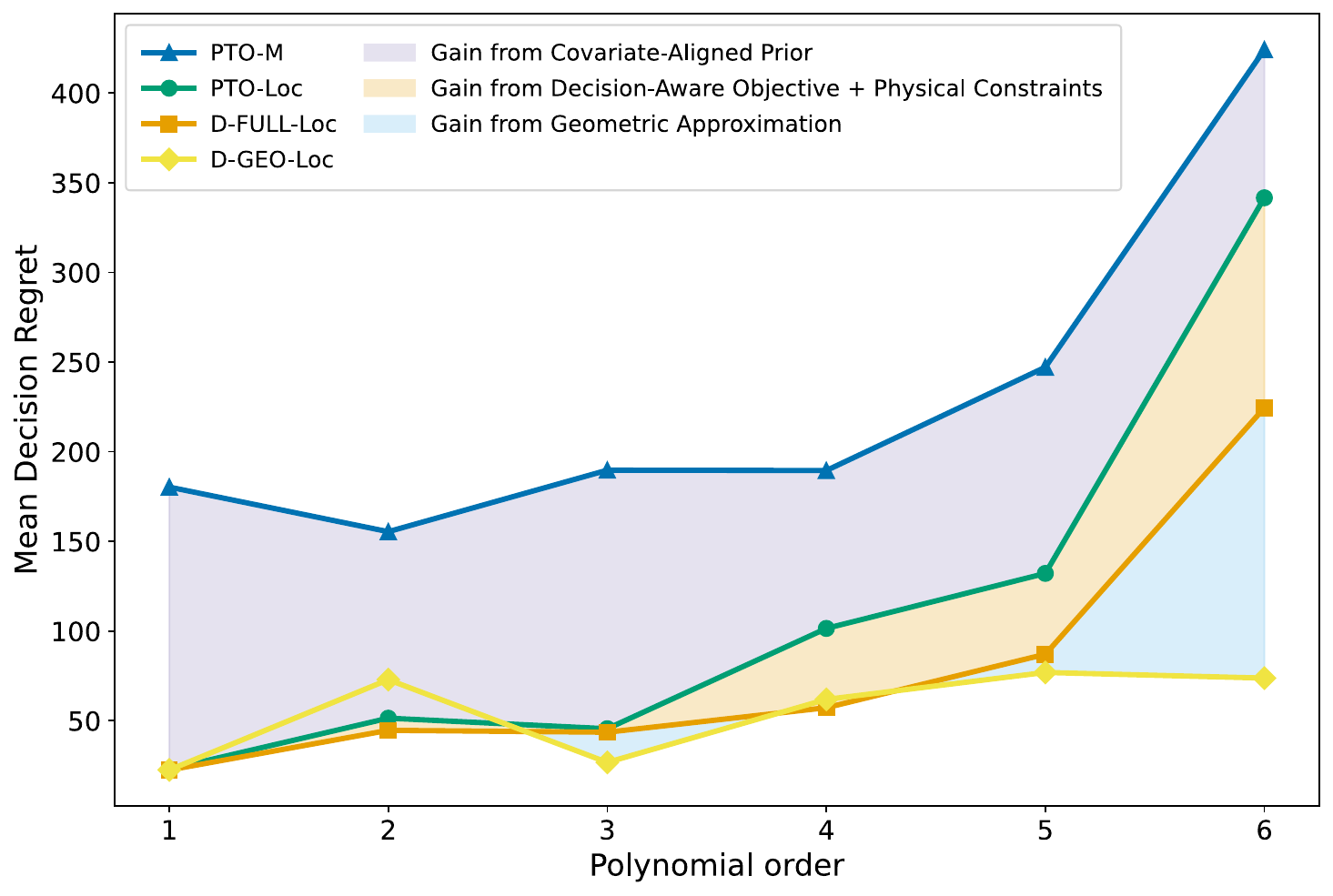}
    \caption{Regret reduction from local prior and decision aware refinement across polynomial orders.}
    \label{fig:SN-Order-Decomp}
\end{figure}

Finally, Figure~\ref{fig:SN-Order-Decomp} decomposes the reduction in regret into three incremental sources. The first is the gain from using a covariate-aligned historical prior. Moving from $\textsc{PTO-M}$ to $\textsc{PTO-Loc}$ lowers regret across all polynomial orders, indicating that
matching historical OD information to the current covariate realization is valuable. However, this improvement alone is not sufficient: in the nonlinear regime, $\textsc{PTO-Loc}$ still exhibits substantial regret. The second gain comes from adding the decision aware objective and physical OD constraints. Relative to $\textsc{PTO-Loc}$, $\textsc{D-Full-Loc}$ further reduces regret, showing that decision aware refinement provides value beyond prediction-centered use of the same prior. The third and largest high-order gain
comes from the geometric approximation. Compared with the full-constraint decision aware estimator, $\textsc{D-Geo-Loc}$ achieves a much lower regret in the nonlinear regime, suggesting that restricting the update to a stable, decision-relevant region is crucial.

Overall, the results provide consistent answers to the three empirical questions posed above. First, increasing level of model misspecification (indicated by demand-generation nonlinearity) does undermine the downstream reliability of prediction-centered methods. Although historical priors that exploit current-instance covariate information improve substantially upon a crude global mean, PTO methods still deteriorate markedly in the highly
nonlinear regime. Second, conditional on using the same historical demand information, decision aware estimation provides additional regret reduction
beyond standard PTO, showing that downstream decision structure contains useful information for OD estimation. However, direct decision aware fitting over the full feasible set can remain unstable when the demand-generation process becomes
sufficiently complex. Third, the benefits of introducing the geometric approximation in the decision aware estimation problem yields clear benefits, reducing both the average and upper-tail regret across training samples.

\subsubsection{Robustness to Noisy Historical OD Labels}

The previous experiments varied the structural complexity of the OD generation process. We next examine a complementary source of misspecification: noisy historical OD labels. The central question is: holding the data-generating process fixed, how sensitive are prediction centered and decision centered estimators to increasing noise in the historical OD information used to construct their priors?


\begin{figure}[t]
    \centering
    \includegraphics[width=0.95\textwidth]{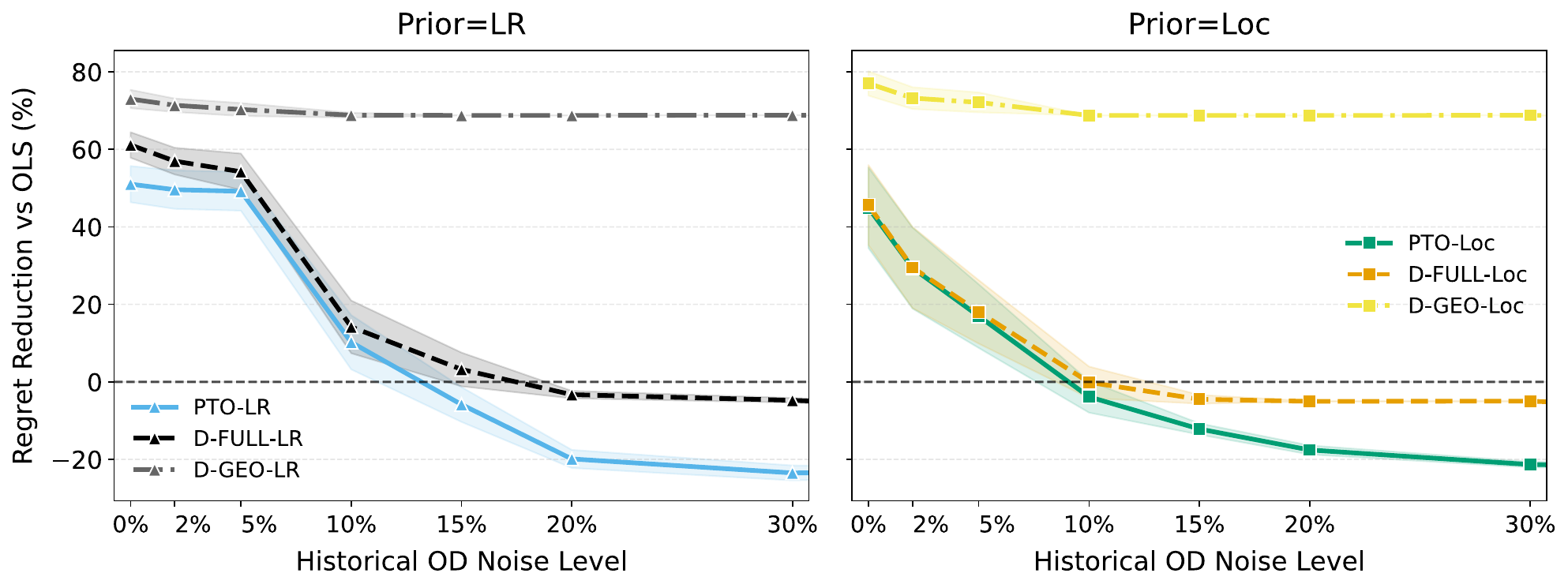}
    \caption{Percentage reduction in decision regret relative to OLS under increasing historical OD label noise.}
    \label{fig:SN-Noise}
\end{figure}

Figure~\ref{fig:SN-Noise} reports the resulting regret sensitivity. In this experiment, we fix the polynomial order at three and add independent nonnegative uniform perturbations to the historical OD demand. The perturbation magnitude increases with the noise level, and the noisy OD entries are clipped to remain within the feasible demand range. The test set and the underlying data-generating process are held fixed, so the experiment isolates the effect of corrupted historical outcome information on downstream decision quality. Since the OLS benchmark is not affected by this historical-label perturbation, we use it as the normalization baseline and report
$$
\text{Reduction vs. OLS}=
100 \cdot \frac{
\text{Regret}_{\mathrm{OLS}}-\text{Regret}_{m}}{
\text{Regret}_{\mathrm{OLS}}}.
$$
The global mean historical prior is omitted because its performance is uniformly poor, allowing the comparison to focus on the $\mathrm{LR}$ and $\mathrm{Loc}$. Under both prior constructions, the $\textsc{PTO}$ methods deteriorate as the historical OD labels become noisier. Both $\textsc{PTO-LR}$ and $\textsc{PTO-Loc}$ initially achieve substantial regret reductions relative to $\textsc{OLS}$, but their advantages vanish and eventually become negative as the noise level increases. The two $\textsc{PTO}$ variants, however, exhibit different deterioration patterns.
$\textsc{PTO-Loc}$ declines more quickly and more smoothly, indicating that the local prior is immediately affected by noisy neighboring historical observations. $\textsc{PTO-LR}$ is more stable at moderate noise levels, but after the perturbation exceeds a critical range, its regret reduction drops sharply and falls below the $\textsc{OLS}$ baseline. This behavior indicates that $\textsc{PTO}$ is highly sensitive to corrupted historical labels, because the downstream optimization is driven by a single predictive reference that becomes increasingly unreliable.

The decision aware estimator with the full structural constraints also becomes less stable under noise. In both panels, $\textsc{D-FULL}$ starts with a sizable reduction relative to OLS but declines sharply as the perturbation level increases. Nevertheless, it continues to improve upon the corresponding $\textsc{PTO}$ baseline over most of the noise range. This suggests that incorporating the decision objective remains beneficial, but an unrestricted decision aware fit can overreact to noisy historical signals. By contrast, the ellipsoidal approximation is substantially more robust. The $\textsc{D-Geo-}r$ methods maintain a large and nearly flat regret reduction across the entire noise range in both the $\mathrm{LR}$ and $\mathrm{Loc}$ panels. Although their performance also decreases mildly as noise increases, the decline is much smaller than that of $\textsc{PTO}$ or $\textsc{D-FULL}$. This stability suggests that the ellipsoidal restriction filters out noisy directions in the historical OD information while preserving the decision-relevant region of the feasible set.

Overall, this experiment reinforces the main robustness message of the paper. Historical OD information is useful only if it is incorporated in a way that is stable under label corruption. Methods that prioritize data alignment alone quickly lose their downstream advantage as historical labels become noisy, whereas decision aware methods with geometric regularization remain robust, preserving meaningful regret reductions even under high noise.

\subsubsection{Sensitivity of the Tuning Parameter $\alpha$ to Label Noise}

We next examine how sensitive the decision aware estimator is to the SPO+ surrogate parameter $\alpha$ when the historical OD prior are noisy. Following the same noisy setting as above, we perturb the historical OD samples used to construct the local prior and evaluate the test-set decision regret of $\textsc{D-Geo-Loc}$ over a grid of positive $\alpha$ values. For each noise level $\eta$, let $R_{\eta}(\alpha)$ denote the average test-set regret obtained using tuning parameter $\alpha$. To measure sensitivity relative to the best value in the grid, we define the excess-regret sensitivity function
$$
S_{\eta}(\alpha)=R_{\eta}(\alpha)-\min_{\alpha' \in \mathcal{A}} R_{\eta}(\alpha'),
$$
where $\mathcal{A}$ is the set of positive $\alpha$ values considered in the experiment. A smaller value of $S_{\eta}(\alpha)$ indicates that the performance at $\alpha$ is close to the best achievable performance for that noise level.


\begin{figure}[t]
    \centering
    \includegraphics[width=0.95\textwidth]{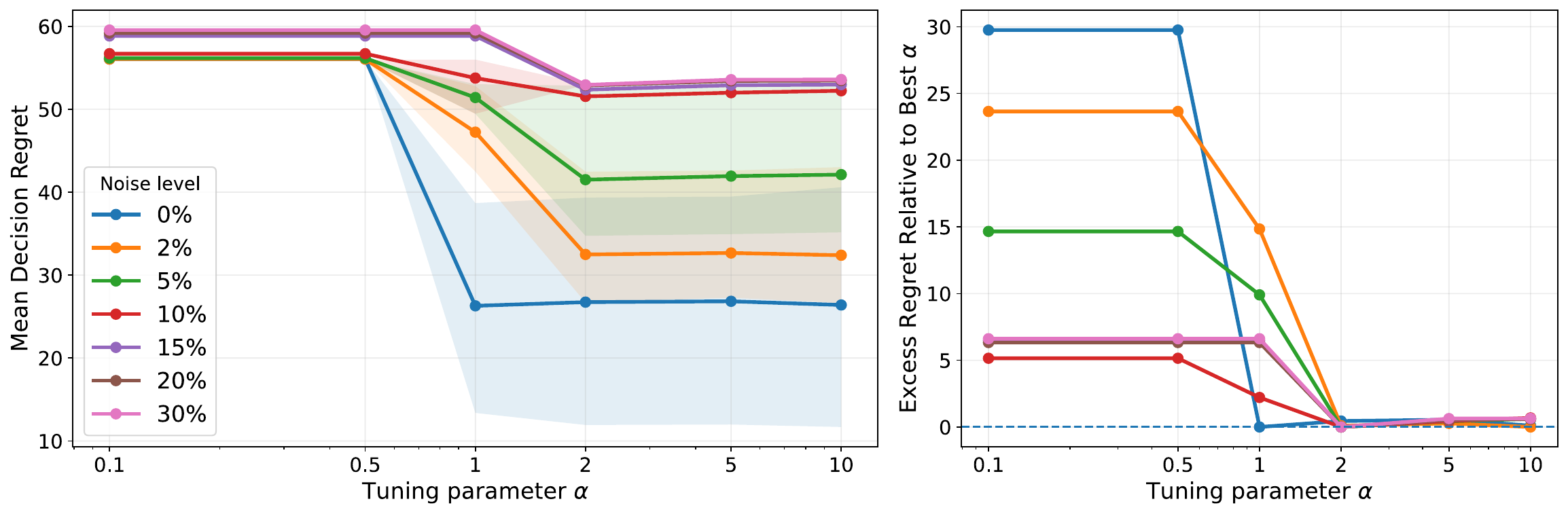}
    \caption{Sensitivity of test-set decision regret to the tuning parameter $\alpha$ under noisy historical OD labels.}
    \label{fig:SN-Alpha}
\end{figure}

Figure~\ref{fig:SN-Alpha} reports the sensitivity of $\textsc{D-Geo-Loc}$ to the tuning parameter $\alpha$ under different levels of label noise. The left panel shows a clear decreasing pattern in average regret as $\alpha$ moves away from very small values. When
$\alpha=0.1$ or $0.5$, the regret remains relatively high across all noise levels, indicating that too small a value of $\alpha$ provides insufficient decision guidance. As $\alpha$ increases to a moderate range, especially around $\alpha=2$, the regret drops substantially and then becomes relatively flat for larger values. This empirical pattern is broadly consistent with the standard SPO+ formulation, where $\alpha=2$ is the canonical scaling used in the Fisher-consistency analysis of the SPO+ surrogate \citep{elmachtoub2022smart}. The right panel reports the excess-regret sensitivity function $S_{\eta}(\alpha)$. Consistent with the left panel, small values of $\alpha$ incur a noticeable excess regret, while moderate and large values produce performance close to the best value within each noise level. This suggests that DWE does not require fine tuning of $\alpha$ under mild to moderate label noise, which is practically useful because the method does not depend on a highly delicate hyperparameter choice. Rather, the main requirement is to avoid overly small values of $\alpha$; once $\alpha$ is sufficiently positive, the estimator achieves relatively stable decision performance across the noise levels considered.


\subsection{Sioux Falls network}\label{EX3}

We further evaluate the proposed decision aware estimation framework on a
larger and more realistic transportation network, the Sioux Falls network.
This network consists of 24 nodes, 76 directed links, and 552 OD pairs. The physical relationship between OD flows and link flows is encoded
by a fixed link-OD incidence matrix $X \in \mathbb{R}^{76 \times 552}$ in
$L = X f$, leading to a substantially higher-dimensional and more severely
underdetermined estimation problem than the synthetic intersection network
considered earlier.

\subsubsection{OD Demand Patterns}


\begin{figure}[t]
    \centering
    \includegraphics[width=0.6\textwidth]{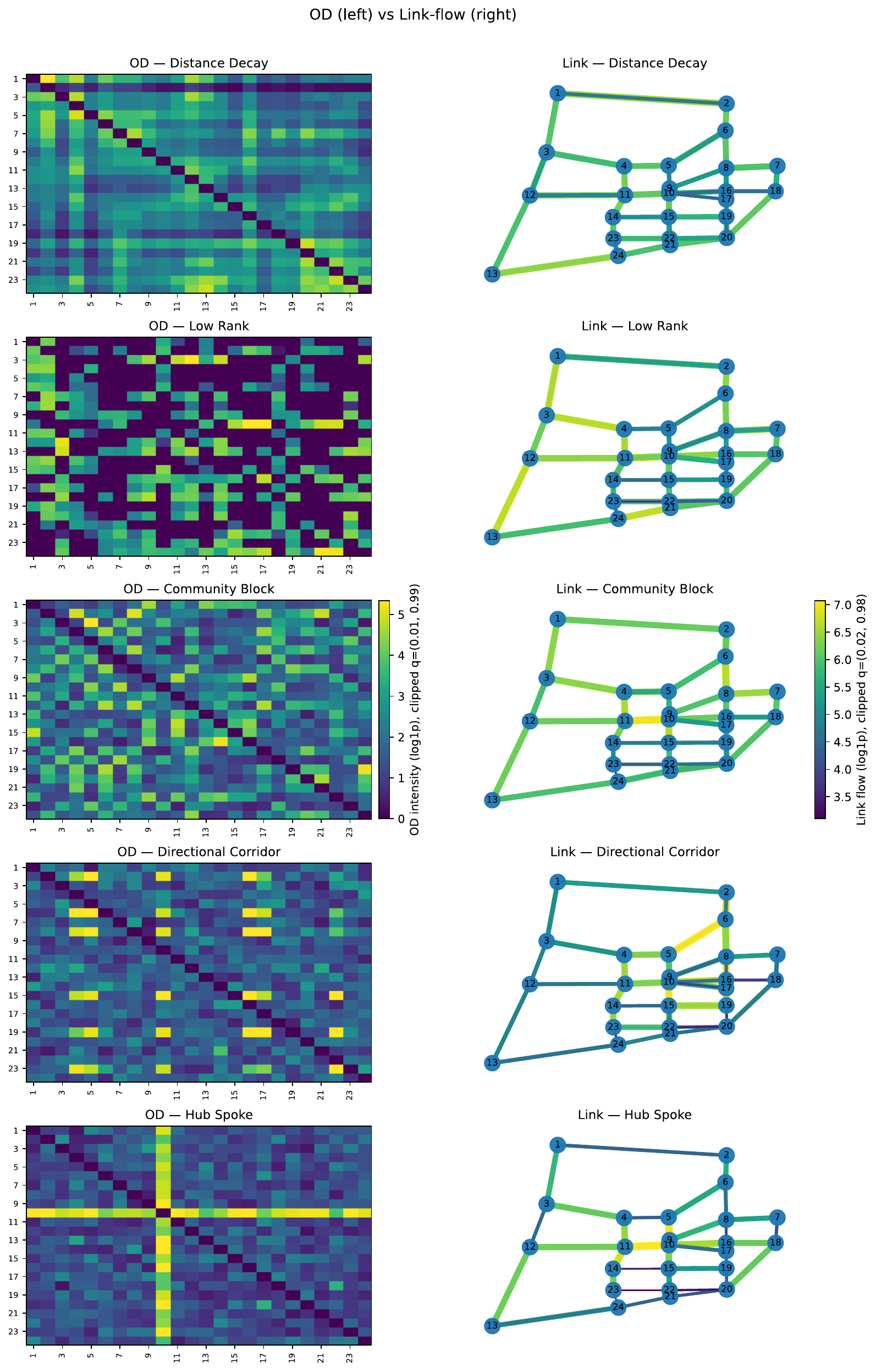}
    \caption{OD patterns (left) and corresponding link flows (right).}
    \label{fig:siouxfalls-patterns}
\end{figure}

To assess robustness across heterogeneous demand structures, we construct five
distinct OD demand patterns on the Sioux Falls network. These patterns are
designed to capture qualitatively different spatial and structural
characteristics commonly observed in real transportation systems: distance
decay, hub-spoke concentration, community block structure, directional
corridors, and low-rank latent demand. Figure~\ref{fig:siouxfalls-patterns}
illustrates the resulting OD matrices together with their induced link-flow
patterns. While these demand modes exhibit markedly different spatial
organizations, all satisfy the same physical flow conservation constraints.
Detailed descriptions of the demand construction and their statistical
properties are provided in Appendix~\ref{app:od_patterns}.

\subsubsection{Experimental Design}

For all five demand patterns, we generate OD-flow priors using a gravity model with imperfect features, providing a common prior across all experiments. We consider three total-demand levels: low ($1{,}000$), medium ($10{,}000$), and high ($20{,}000$).  


We compare OLS, PTO, and two DWE variants. OLS reconstructs OD flows from observed link flows using least squares, while PTO directly optimizes the downstream problem using the gravity-model prediction. Following the previous name convention,  DWE methods are denoted by D-Full-$\mathrm{G}$ and D-Geo-$\mathrm{G}$, where $\mathrm{G}$ denotes the gravity-model prior. D-Full-G directly solves the full decision aware estimation formulation, whereas D-Geo-G uses the proposed geometry-based formulation; for D-Geo-G, we evaluate multiple radii and report the best-performing radius $r^*$.

Table~\ref{tab:demand_performance} summarizes the performances across all demand patterns and demand levels, evaluated using downstream decision regret.

\begin{table}[htbp]
\centering
\caption{Performance across demand levels and demand modes.}
\begin{threeparttable}

\label{tab:demand_performance}

\small
\setlength{\tabcolsep}{6pt}
\renewcommand{\arraystretch}{1.15}

\begin{tabular}{llrrrrr}
\toprule

\multirow{2}{*}{\textbf{Demand Level}}
& \multirow{2}{*}{\textbf{Demand Mode}}
& \multirow{2}{*}{\textbf{RMSE}}
& \multirow{2}{*}{\textbf{OLS}}
& \multirow{2}{*}{\textbf{PTO}}
& \multicolumn{2}{c}{\textbf{DWE Variants}} \\

\cmidrule(lr){6-7}

&
&
&
&
&
\textbf{D-FULL-G}
& \textbf{D-GEO-G} \\

\midrule

\multirow{5}{*}{Low}
& Distance Decay
& 0.55
& 25.77
& \textbf{2.59}
& \textbf{2.59}
& \textbf{2.59} $(r^*=5)$ \\

& Low Rank
& 1.74
& 59.84
& 12.74
& \textbf{3.99}
& \textbf{3.99} $(r^*=10)$ \\

& Community Blocks
& 2.57
& 55.15
& 42.18
& \textbf{5.45}
& \textbf{5.45} $(r^*=10)$ \\

& Directional Corridor
& 1.59
& 52.97
& 26.91
& 6.09
& \textbf{0.00} $(r^*=3)$ \\

& Hub Spoke
& 1.75
& 92.92
& 3.77
& \textbf{0.15}
& \textbf{0.15} $(r^*=10)$ \\

\midrule

\multirow{5}{*}{Medium}
& Distance Decay
& 5.33
& 320.14
& 36.19
& 36.19
& \textbf{3.36} $(r^*=3)$ \\

& Low Rank
& 17.11
& 795.97
& 257.26
& \textbf{12.81}
& \textbf{12.81} $(r^*=10)$ \\

& Community Blocks
& 25.13
& 560.63
& 430.56
& 73.84
& \textbf{28.59} $(r^*=5)$ \\

& Directional Corridor
& 16.14
& 513.44
& 198.08
& 94.74
& \textbf{0.00} $(r^*=3)$ \\

& Hub Spoke
& 17.55
& 1018.62
& 45.33
& \textbf{0.00}
& \textbf{0.00} $(r^*=10)$ \\

\midrule

\multirow{5}{*}{High}
& Distance Decay
& 11.42
& 228.24
& \textbf{3.84}
& \textbf{3.84}
& \textbf{3.84} $(r^*=10)$ \\

& Low Rank
& 35.78
& 1420.49
& \textbf{195.95}
& \textbf{195.95}
& \textbf{195.95} $(r^*=5)$ \\

& Community Blocks
& 52.50
& 949.19
& 873.83
& \textbf{271.38}
& \textbf{271.38} $(r^*=10)$ \\

& Directional Corridor
& 31.52
& 1067.51
& 585.95
& 214.22
& \textbf{0.00} $(r^*=3)$ \\

& Hub Spoke
& 34.93
& 1609.66
& \textbf{44.75}
& \textbf{44.75}
& \textbf{44.75} $(r^*=10)$ \\

\bottomrule
\end{tabular}

\begin{tablenotes}[flushleft]
\footnotesize
\item[] \textit{Notes:}
The RMSE column reports the prediction RMSE of the prior baseline.
All OLS, PTO, and DWE columns report  decision regret divided by $10^3$.
For D-GEO-G, the best result over all tested radii is reported, with the selected radius shown as $r^*$.
Bold values indicate the best decision regret within each setting.
\end{tablenotes}

\end{threeparttable}
\end{table}

\subsubsection{Results}

Table~\ref{tab:demand_performance} documents a clear advantage of decision aware estimation on the Sioux Falls network. OLS yields the largest decision regret in all 15 demand-level--mode combinations, highlighting the limitations of recovering OD demand from link-flow consistency alone in a severely underdetermined system. PTO substantially improves upon OLS by incorporating the gravity-model prediction, but considerable decision regret remains in several settings. Both DWE variants deliver further gains: D-Full-G achieves strictly lower decision regret than PTO in 10 of the 15
cases and matches it in the remaining five, while D-Geo-G improves upon PTO in 11 cases and matches it in the other four. Among the cases with strict improvement, the reduction relative to PTO ranges from $52.2\%$ to $100\%$ for D-Full-G and from $68.7\%$ to $100\%$ for D-Geo-G. Across all 15 cases, the simple average regret decreases from $184.00$ under PTO to $64.40$ under D-Full-G and $38.19$ under D-Geo-G, corresponding to reductions of approximately $65.0\%$ and $79.2\%$, respectively. To assess whether these conclusions are robust to variation across experimental realizations, we conduct an additional repeated-sampling experiment reported in Appendix Table~\ref{tab:robust}. Each setting is repeated 20 times, and the mean decision regret is reported. D-Full-G and D-Geo-G strictly outperform PTO and OLS in all 15 settings, confirming the robustness of decision aware estimation. Thus, explicitly accounting for the downstream decision produces substantial improvements without deteriorating decision quality in any of the baseline scenarios.  A direct comparison of the two DWE variants provides evidence of an additional regularization effect from the geometric formulation. For D-Geo-G, Table~\ref{tab:demand_performance} reports the best result over the tested radii. D-Geo-G achieves decision regret that is lower than or equal to that of D-Full-G in all 15 cases: it performs strictly better in five cases and matches D-Full-G in the remaining ten. Its average regret is also
$40.7\%$ lower than that of D-Full-G ($38.19$ versus $64.40$). 

One may notice that the magnitude of decision regrets varies sharply across demand patterns.  It would be interesting to explore how demand-prediction error propagates through downstream problems of different network structures, but it is beyond the scope of this work.

\begin{figure}[t]
    \centering
    \includegraphics[width=\textwidth]{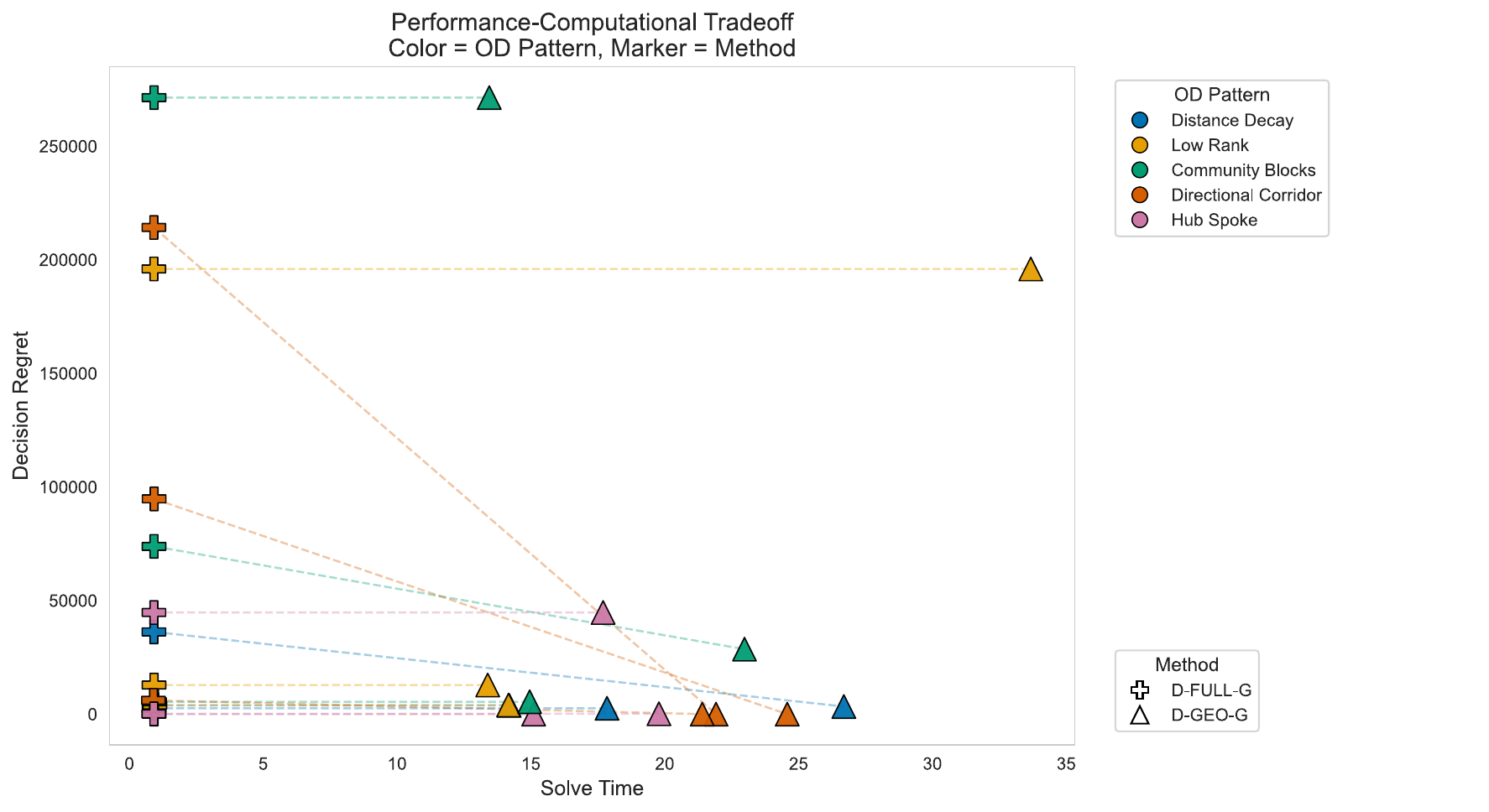}
    \caption{Performance--computational trade-off across different OD patterns and solution methods, measured by decision regret and solve time.}
    \label{fig:performance-computational-tradeoff}
\end{figure}

On the other hand, the gain of D-Geo-G in performance robustness comes at a substantial computational cost. As shown in Figure~\ref{fig:performance-computational-tradeoff}, D-Full-G solves in $0.922$ seconds on average, with a median of $0.921$ seconds and a standard deviation of only $0.011$ seconds. D-Geo-G, by contrast, requires $19.455$ seconds on average, with a median of $17.845$ seconds and a standard
deviation of $5.835$ seconds. Thus, D-Geo-G requires approximately $21.1$ times the mean solve time of D-Full-G. The runtime of D-Full-G across all instances is similar, while D-Geo-G`s runtime seem to vary by instances.  D-Full-G is therefore preferable when computational efficiency is paramount, while D-Geo-G offers a more conservative alternative when the additional protection against over-adaptation justifies the computational premium.

\section{Conclusions and Future Research}\label{DAODE-CON}

As we conclude this study, we reflect on a broader lesson that extends beyond OD estimation. OD demand is not directly observed and must instead be inferred from indirect information, such as observed link flows, physical network constraints, and prior demand patterns. These sources help characterize which OD estimates are plausible, but plausibility alone does not determine which estimate is most useful. A key premise we emphasize is: \textit{not all estimation errors are consequential in the same way}. A good estimate should be faithful enough to the world we observe, but attentive to the consequences of the decisions it informs.  Through analytical and numerical results, we have shown that accounting for how estimates are ultimately used can substantially improve their value for downstream decision-making.

Beyond knowledge from transportation science, our research design draws on two complementary methodological perspectives: geometry and stochastic optimization. A geometric perspective shapes how we view and represent the demand space. OD demand vectors are typically modeled in a high-dimensional space whose dimension equals the number of OD pairs in the network. A central message from this study is that a realistic OD demand does not occupy the entire ambient space. Instead, it is implicitly restricted by a variety of physical and behavioral constraints. These structural restrictions imply that feasible OD demand vectors lie on a much lower-dimensional manifold embedded in the high-dimensional OD space. Ignoring this geometric structure can lead to misleading conclusions in data-driven prediction. A stochastic optimization perspective provides a flexible framework for integrating data, prior knowledge, physical constraints, and decision regret within a unified model. Equally important, advances in stochastic optimization provide the theoretical and computational tools needed to solve the resulting models efficiently and reliably.  Our experience with these complementary perspectives also points to a broader lesson: new research challenges often benefit from ideas that lie outside the traditional boundaries of a discipline. Particularly for young scholars, the broader the set of intellectual building blocks one acquires, the better prepared one may be to construct solutions to challenges not yet anticipated.  


The present study opens several promising directions for future research, spanning modeling generality, data reliability, theoretical analysis, and algorithmic development. First and foremost, while the linear constraints considered in this study capture fundamental flow conservation laws, real transportation systems are governed by richer physical mechanisms, including nonlinear constraints and dynamic relations that may be expressed through differential or difference equations. Accommodating such general physical constraints would substantially broaden its applicability, while also posing new analytical and computational challenges.  Second, the reliability of link-flow observations themselves deserves explicit attention. In this paper, link flows are treated as accurate measurements that define a consistent constraint system. Developing principled methods to handle imperfect link-flow observations, such as through soft constraints, feasibility robust formulations, or joint estimation of OD flows and link-flow errors, remains an important and challenging direction for future research.  In additional, a natural extension is to consider more complex decision problems, including those with nonlinear objectives, integer decisions, equilibrium constraints, or multistage structures. Such extensions would allow the framework to capture a broader class of transportation decisions, but would also require new tools to analyze regret, tractability, and solution stability. Finally, how to characterize regret presents future research opportunities.  For example, while the current results provide initial bounds, sharper regret guarantees—such as tighter upper bounds, lower bounds, or instance-dependent characterizations—are needed to better understand the fundamental limits of decision aware OD estimation.  These directions are not simply extensions of the present framework; they point toward a broader opportunity to rethink how estimation and decision-making should interact. As increasingly rich data and increasingly consequential decisions become intertwined, we believe that learning not only from the data, but also from how the data will ultimately be used, offers a promising path forward.

 \ACKNOWLEDGMENT{This work is supported in part by the National Science Foundation under grant CMMI-2432336/CMMI-2432337.}

\bibliographystyle{informs2014trsc} 
\bibliography{references}

\begin{APPENDICES}

\section{Ellipsoidal Radius and Calibration}
\label{app:ellipsoid-scale}

Let $d=\dim(\ker X)=n-q$ and consider the ellipsoidal search region $\mathcal B_H(r)=\left\{u\in\mathbb R^d
\;\middle|\;u^\top H u\le r^2\right\}$, where $H\succ0$ is the Hessian of the logarithmic barrier evaluated at
the analytic-center. Let $H=Q\Lambda Q^\top$, $\Lambda=\operatorname{diag}(\lambda_1,\ldots,\lambda_d)$. Under the coordinate transformation $y = Q^\top u$, the ellipsoid becomes $\sum_{i=1}^d \lambda_i y_i^2 \le r^2$, or equivalently $\sum_{i=1}^d y_i^2/(r^2/\lambda_i) \le 1$. Comparing this with the standard ellipsoid form $\sum_{i=1}^d y_i^2/a_i^2 \le 1$, the principal semiaxis along the $i$th eigenvector of $H$ is $a_i(r)=r/\sqrt{\lambda_i}$. Thus, $r$ controls the overall scale of the ellipsoid, while the eigenvalues $\lambda_i$ determine its relative shape across directions. Increasing $r$ by a factor $\alpha$ expands every principal semiaxis by the same factor $\alpha$.

The effect of dimension can also be seen from the volume of the ellipsoid. Its $d$-dimensional volume is
$$
\operatorname{Vol}_d\bigl(\mathcal B_H(r)\bigr)
=
\frac{\pi^{d/2}}{\Gamma(d/2+1)}
\frac{r^d}{\sqrt{\det H}}.
$$
Defining $g_H=(\det H)^{1/d}$, and applying Stirling's approximation gives
$$
\operatorname{Vol}_d\bigl(\mathcal B_H(r)\bigr)
\sim
\frac{1}{\sqrt{\pi d}}
\left(
\frac{2\pi e\,r^2}{d\,g_H}
\right)^{d/2}.
$$
Therefore, under a fixed coordinate normalization, the volume represented
by a fixed radius $r$ can change substantially with the nullspace
dimension $d$. For example, when $H=I$ and $r=1$, the ellipsoid reduces to the Euclidean unit ball. Its volume decreases from approximately $2.55$ in $10$ dimensions to $2.58\times10^{-2}$ in $20$ dimensions, and further to $1.73\times10^{-13}$ in $50$ dimensions. Thus, the same numerical value of $r$ can correspond to very different
volumetric coverage as the dimension of the nullspace changes.

This observation suggests a simple heuristic for choosing the radius as the dimension changes. From the leading term in the volume approximation, the quantity $\frac{r^2}{d\,g_H}$ determines the exponential scale of the ellipsoid volume, suggesting the scaling $r\propto\sqrt{d\,g_H}$. We first select a representative reference instance and choose $r_0$ from a set of candidate radii by minimizing the downstream regret on a validation set. Let $d_0$ and $H_0$ denote the nullspace dimension and Hessian of this reference instance, and let $g_{H,0}=(\det H_0)^{1/d_0}$ denote the corresponding Hessian geometric mean. We scale the validated radius $r_0$ to a new instance according to
$$
r_{\mathrm{heur}}=\max\left\{
1,\;
r_0
\sqrt{
\frac{d\,g_H}
     {d_0\,g_{H,0}}
}
\right\}.
$$

For example, if $g_H\approx g_{H,0}$ and validation selects $r_0=1$ for a reference instance with $d_0=10$, the rule gives approximately $r_{\mathrm{heur}}=2$ for $d=40$ and $r_{\mathrm{heur}}=3$ for $d=90$. In this case, the rule reduces to $r\propto\sqrt{d}$. The heuristic transfers the radius calibrated on the reference instance to problems with different dimensions and Hessian geometries, and the resulting value can be further refined using validation data if needed.

\section{Construction of the Local Error Distribution}
\label{app:local-neighborhood}

This appendix describes how the error distribution $\mathcal{P}$ can be constructed from historical data and contextual information. Let
$\mathcal{D}=\{(x_d,r_d)\}_{d=1}^D$ denote a historical dataset, where $x_d$ contains contextual information associated with observation $d$, such as historical demand, socioeconomic characteristics, land use, and other factors related to travel demand. The residual $r_d$ measures the error of the prior OD estimate for that observation. In particular, if $f_{0,d}$ denotes the prior estimate and $f_d^{\mathrm{obs}}$ the corresponding historical OD observation, we define
$r_d=f_d^{\mathrm{obs}}-f_{0,d}$.

For a target instance with contextual features $x^\star$, we use historical observations with similar features to characterize the uncertainty in the prior estimate $f_0$. Following the local weighting perspective of \cite{bertsimas2020predictive}, we define a neighborhood $\mathcal{N}(x^\star)\subseteq\{1,\ldots,D\}$ that determines which historical residuals contribute to the local error distribution. A simple empirical construction is
$$
\widehat{\mathcal{P}}(x^\star)=\frac{1}{|\mathcal{N}(x^\star)|}
\sum_{d\in\mathcal{N}(x^\star)}
\delta_{r_d},
$$
where $\delta_{r_d}$ denotes a point mass at residual $r_d$. Different methods for defining $\mathcal{N}(x^\star)$ correspond to different notions of similarity in the feature space. We consider several standard constructions below.

\subsection{$k$-nearest neighbors (kNN)}

The $k$-nearest-neighbor neighborhood consists of the $k$ historical observations whose feature vectors are closest to $x^\star$ under a chosen distance metric $dis(\cdot,\cdot)$:
\begin{equation}
\mathcal{N}_{\mathrm{kNN}}(x^\star)
=
\left\{
d : x_d \text{ is among the } k \text{ closest points to } x^\star
\text{ under } dis(\cdot,\cdot)
\right\}.
\end{equation}
This construction fixes the number of residual observations used to estimate the local error distribution.

\subsection{$\varepsilon$-ball neighborhood}

An $\varepsilon$-ball neighborhood includes all historical observations within distance $\varepsilon$ of $x^\star$:
\begin{equation}
\mathcal{N}_{\varepsilon}(x^\star)
=
\left\{
d : \|x_d-x^\star\|\le\varepsilon
\right\}.
\end{equation}
The resulting neighborhood size adapts to the local density of the feature space.

\subsection{Decision tree-based neighborhood}

A decision tree trained on the historical data partitions the feature space into axis-aligned regions. The neighborhood contains observations that fall in the same terminal leaf as $x^\star$:
\begin{equation}
\mathcal{N}_{\mathrm{tree}}(x^\star)
=
\left\{
d : x_d \text{ and } x^\star
\text{ lie in the same terminal leaf}
\right\}.
\end{equation}
This construction allows the neighborhood to reflect feature relevance and interactions learned from the historical data.

\subsection{Kernel-based neighborhood}

A kernel-based construction assigns each historical residual a weight
\begin{equation}
w_d
=
K\!\left(
\frac{\|x_d-x^\star\|}{h}
\right),
\end{equation}
where $K(\cdot)$ is a kernel function and $h$ is a bandwidth parameter. After normalization, $\tilde w_d=w_d/\sum_{j=1}^D w_j$, the local error distribution can be written as
$$
\widehat{\mathcal{P}}(x^\star)
=
\sum_{d=1}^D
\tilde w_d\,\delta_{r_d}.
$$
This weighted construction avoids a discrete neighborhood boundary and gives greater influence to historical observations that are more similar to the target instance.

\section{Proof of Proposition~\ref{prop:spo_plus_ours}}
\label{APP:PP2}

We prove the three stated properties of $\ell_{\alpha}$ for any realized OD-flow vector $f^\ast$, candidate OD estimate $f$, and $\alpha>0$.

\proof{Proof.}
We begin by recalling the definition of regret. For a given decision rule $w(f)$, the regret associated with a realized OD flow $f^\ast$ is
$$
R(f^\ast,f)={f^\ast}^{\!\top} w(f) - c(f^\ast),
$$
where $c(f^\ast):=\min_{w\in\mathcal{W}} {f^\ast}^{\!\top} w={f^\ast}^{\!\top} w(f^\ast)$ and $w(f)\in\arg\min_{w\in\mathcal{W}} f^\top w$ denotes an optimal solution to the downstream optimization problem under $f$.

\medskip
\noindent\emph{Step 1 (Upper bound).}
Fix any $\alpha>0$. Using $f^\top w(f)=c(f)$, we have
$$
\begin{aligned}
R(f^\ast,f)
&=
{f^\ast}^{\!\top}w(f)-c(f^\ast)\\
&=
(f^\ast-\alpha f)^\top w(f)
+\alpha c(f)-c(f^\ast)\\
&\le
\max_{w\in\mathcal W}
(f^\ast-\alpha f)^\top w
+\alpha c(f)-c(f^\ast).
\end{aligned}
$$
Since $w(f^\ast)\in\mathcal W$ is feasible for the downstream problem under $f$,
$$
c(f)=f^\top w(f)
\le
f^\top w(f^\ast).
$$
Because $\alpha>0$, it follows that
$$
\begin{aligned}
R(f^\ast,f)
&\le
\max_{w\in\mathcal W}
(f^\ast-\alpha f)^\top w
+
\alpha f^\top w(f^\ast)
-
c(f^\ast)\\
&=
\ell_\alpha(f^\ast,f).
\end{aligned}
$$
Hence,
$$
R(f^\ast,f)\le \ell_\alpha(f^\ast,f),
$$
which proves the first property in Proposition~\ref{prop:spo_plus_ours}.

\medskip
\noindent\emph{Step 2 (Convexity).}
For fixed $f^\ast$, the surrogate can be written as
$$
\ell_\alpha(f^\ast,f)
=
\max_{w\in\mathcal W}
\left\{
(f^\ast-\alpha f)^\top w
+
\alpha f^\top w(f^\ast)
-
c(f^\ast)
\right\}.
$$
For each fixed $w\in\mathcal W$, the expression inside the maximum is affine, and hence convex, in $f$. Therefore, by the supremum rule for convex functions
\cite[Proposition~1.18(a)]{royset2021optimization}, $\ell_\alpha(f^\ast,\cdot)$ is convex. This proves the second property in Proposition~\ref{prop:spo_plus_ours}.

\medskip
\noindent\emph{Step 3 (Subgradient characterization).}

For fixed $f^\ast$ and each $w\in\mathcal W$, the expression inside the maximum in~\eqref{eq:spo_plus_ours} is affine in $f$, with gradient $\alpha\bigl(w(f^\ast)-w\bigr)$. Since $\mathcal W$ is compact, Proposition~6.30 of
\cite{royset2021optimization} implies
$$
\partial_f\ell_\alpha(f^\ast,f)=\operatorname{con}
\left\{\alpha\bigl(w(f^\ast)-w\bigr):w\in
\arg\max_{\tilde w\in\mathcal W}
(f^\ast-\alpha f)^\top\tilde w
\right\},
$$
where $\operatorname{con}(\cdot)$ denotes the convex hull. Moreover,
$$
\arg\max_{w\in\mathcal W}
(f^\ast-\alpha f)^\top w=\arg\min_{w\in\mathcal W}(\alpha f-f^\ast)^\top w.
$$
Therefore, choosing
$w(\alpha f-f^\ast)$ from this optimal set yields
$$
\alpha
\bigl(
w(f^\ast)-w(\alpha f-f^\ast)
\bigr)
\in
\partial_f\ell_\alpha(f^\ast,f),
$$
which proves the third property in
Proposition~\ref{prop:spo_plus_ours}.
\hfill$\square$
\endproof

\section{Proof of Proposition \ref{prop:alpha-tradeoff} }
\label{app:Pro2}

\proof{Proof.}

We first decompose the discrepancy between the surrogate evaluated at the
realized OD flow $f^\ast$ and the regret under the true OD flow
$f_{\rm true}$ as
$$
\ell_\alpha(f^\ast,f)-R(f_{\rm true},f)
=\bigl[\ell_\alpha(f_{\rm true},f)-R(f_{\rm true},f)\bigr]+\bigl[\ell_\alpha(f^\ast,f)-\ell_\alpha(f_{\rm true},f)\bigr].
$$
We bound these two terms separately.

\medskip
\noindent\emph{Step 1 (Surrogate approximation).}
From the definitions of $\ell_\alpha$ and $R$, we have
$$
\begin{aligned}
\ell_\alpha(f_{\rm true},f)-R(f_{\rm true},f)
={}&
\max_{w\in\mathcal W}
(f_{\rm true}-\alpha f)^\top
\bigl(w-w(f)\bigr)+
\alpha f^\top
\bigl(w(f_{\rm true})-w(f)\bigr).
\end{aligned}
$$

For any $w\in\mathcal W$, the boundedness of $f_{\rm true}$ and the sharpness condition imply
$$
\begin{aligned}
(f_{\rm true}-\alpha f)^\top\bigl(w-w(f)\bigr)&=f_{\rm true}^\top\bigl(w-w(f)\bigr)-\alpha f^\top\bigl(w-w(f)\bigr)\\
&\le \bigl(U-\alpha\kappa\bigr)\|w-w(f)\|.
\end{aligned}
$$
If $\alpha\kappa\ge U$, the right-hand side is nonpositive for every
$w\in\mathcal W$, while choosing $w=w(f)$ gives zero. Hence, the maximum is
zero. If $\alpha\kappa<U$, then $\|w-w(f)\|\le 2B_w$ and
$U<U^2/(\kappa\alpha)$, so
$$
\max_{w\in\mathcal W}
(f_{\rm true}-\alpha f)^\top
\bigl(w-w(f)\bigr)
\le
\frac{2U^2B_w}{\kappa\alpha}.
$$
Therefore, the same bound holds for every $\alpha>0$.

For the second term, the optimality of $w(f_{\rm true})$ under
$f_{\rm true}$ gives
$f_{\rm true}^\top(w(f_{\rm true})-w(f))\le0$. Thus,
$$
\begin{aligned}
f^\top\bigl(w(f_{\rm true})-w(f)\bigr)
&=
(f-f_{\rm true})^\top
\bigl(w(f_{\rm true})-w(f)\bigr)+
f_{\rm true}^\top
\bigl(w(f_{\rm true})-w(f)\bigr)\\
&\le
2B_w\|f-f_{\rm true}\|.
\end{aligned}
$$
Combining the two bounds yields
$$
\ell_\alpha(f_{\rm true},f)-R(f_{\rm true},f)
\le
\frac{2U^2B_w}{\kappa\alpha}
+
2\alpha B_w\|f-f_{\rm true}\|.
$$

\medskip
\noindent\emph{Step 2 (Uncertainty perturbation).}
Using the definition of $\ell_\alpha$,
$$
\begin{aligned}
\ell_\alpha(f^\ast,f)-\ell_\alpha(f_{\rm true},f)&=
\max_{w\in\mathcal W}(f^\ast-\alpha f)^\top w
-
\max_{w\in\mathcal W}(f_{\rm true}-\alpha f)^\top w\\
&+
\alpha f^\top
\bigl(w(f^\ast)-w(f_{\rm true})\bigr)
+
c(f_{\rm true})-c(f^\ast).
\end{aligned}
$$

The first difference is bounded by
$$
\begin{aligned}
&
\max_{w\in\mathcal W}(f^\ast-\alpha f)^\top w
-
\max_{w\in\mathcal W}(f_{\rm true}-\alpha f)^\top w\le
\max_{w\in\mathcal W}
(f^\ast-f_{\rm true})^\top w
\le
B_w\|f^\ast-f_{\rm true}\|.
\end{aligned}
$$
Similarly, since $w(f^\ast)$ is feasible under $f_{\rm true}$,
$$
c(f_{\rm true})-c(f^\ast)
\le
(f_{\rm true}-f^\ast)^\top w(f^\ast)
\le
B_w\|f^\ast-f_{\rm true}\|.
$$

It remains to bound the term involving the two optimal decisions. Since
$w(f^\ast)$ is optimal under $f^\ast$,
${f^\ast}^\top(w(f^\ast)-w(f_{\rm true}))\le0$. Hence,
$$
\begin{aligned}
f^\top\bigl(w(f^\ast)-w(f_{\rm true})\bigr)=
(f-f^\ast)^\top
\bigl(w(f^\ast)-w(f_{\rm true})\bigr)+
{f^\ast}^\top
\bigl(w(f^\ast)-w(f_{\rm true})\bigr)\le
2B_w\|f^\ast-f\|.
\end{aligned}
$$
Therefore,
$$
\ell_\alpha(f^\ast,f)-\ell_\alpha(f_{\rm true},f)
\le
2\alpha B_w\|f^\ast-f\|
+
2B_w\|f^\ast-f_{\rm true}\|.
$$

\medskip
\noindent\emph{Step 3 (Combining the bounds).}
Combining Steps~1 and~2 gives
$$
\begin{aligned}
\ell_\alpha(f^\ast,f)-R(f_{\rm true},f)
\le\;&
\frac{2U^2B_w}{\kappa\alpha}
+
2\alpha B_w
\left(
\|f-f_{\rm true}\|+\|f^\ast-f\|
\right)+
2B_w\|f^\ast-f_{\rm true}\|,
\end{aligned}
$$
which proves~\eqref{eq:alpha-tradeoff-bound}.

Finally, suppose that
$\|f-f_{\rm true}\|+\|f^\ast-f\|>0$. The part of the bound that depends on
$\alpha$ is
$$
\frac{2U^2B_w}{\kappa\alpha}
+
2\alpha B_w
\left(
\|f-f_{\rm true}\|+\|f^\ast-f\|
\right).
$$
Differentiating with respect to $\alpha$ and setting the derivative equal to
zero gives
$$
-\frac{2U^2B_w}{\kappa\alpha^2}
+
2B_w
\left(
\|f-f_{\rm true}\|+\|f^\ast-f\|
\right)
=0.
$$
Solving for $\alpha$ yields
$$
\alpha^\ast
=
\frac{U}{
\sqrt{
\kappa\left(
\|f-f_{\rm true}\|+\|f^\ast-f\|
\right)
}}.
$$
Since the second derivative is positive for every $\alpha>0$, this is the
unique minimizer of the $\alpha$-dependent part of the bound. This proves
\eqref{eq:oracle-alpha}.
\hfill$\square$
\endproof

\section{Proof of Theorem \ref{T1}}

\proof{Proof.}
For a fixed realization $\varepsilon$, define
$$
h_\varepsilon(z):=
\ell_{\alpha}(f_0+\varepsilon,g(z)),
\qquad
g(z)=\bar f+Vz.
$$
By Proposition~\ref{prop:spo_plus_ours},
$\ell_{\alpha}(f_0+\varepsilon,\cdot)$ is convex. Since $g(z)$ is affine,
the affine composition rule implies that $h_\varepsilon(z)$ is convex in $z$.
Therefore,
$$
\phi(z)=\mathbb{E}_{\varepsilon\sim\mathcal P}\left[h_\varepsilon(z)\right],
$$
is convex, since expectation preserves convexity. Moreover, the feasible set
$\{z:z^\top H z\le r^2\}$ is convex because $H\succ0$. Hence, the expectation
formulation is a convex optimization problem.

We next establish the subgradient expression. By
Proposition~\ref{prop:spo_plus_ours}, for each realization
$f_0+\varepsilon$,
$$
\alpha \left[w(f_0+\varepsilon)-w\bigl(\alpha g(z)-f_0-\varepsilon\bigr)\right]\in \partial_f \ell_{\alpha}(f_0+\varepsilon,g(z)).
$$
Since $g(z)=\bar f+Vz$ is affine, the affine subgradient chain rule gives
$$
\alpha V^\top \left[w(f_0+\varepsilon)-w\bigl(\alpha g(z)-f_0-\varepsilon\bigr)\right]\in \partial_z h_\varepsilon(z).
$$

Under the stated integrability conditions, Proposition~3.23 of
\citet{royset2021optimization} allows the expectation and subgradient
operations to be interchanged. Therefore,
$$
\mathbb{E}_{\varepsilon\sim\mathcal P}
\left[\alpha V^\top \left(w(f_0+\varepsilon)-w\bigl(\alpha g(z)-f_0-\varepsilon\bigr)\right)\right] \in \partial\phi(z),
$$
which is the subgradient expression stated in Theorem~\ref{T1}.
\hfill$\square$
\endproof

\section{Extension to Alternative Risk Measures}
\label{app:cvar_extension}

This appendix provides the formal statement and proof of the extension mentioned
in the main text, showing that the convexity and subgradient characterization
also applies to the Conditional Value-at-Risk (CVaR) formulation. Throughout
this appendix, $\beta\in(0,1)$ denotes the CVaR confidence level, while
$\alpha>0$ remains the scaling parameter in the surrogate loss.

\begin{theorem}[Convexity and Subgradient of the CVaR Formulation]
\label{T2}
Let $H\in\mathbb{R}^{d\times d}$ be positive definite and consider
$$
\min_{z\in\mathbb{R}^d:\;z^\top H z\le r^2}
\psi(z):=\mathrm{CVaR}_{\beta}\left[\ell_{\alpha}(f_0+\varepsilon,g(z))\right],\qquad \varepsilon\sim\mathcal P,
$$
where $g(z)=\bar f+Vz$. Assume that the loss and the subgradient expression
below are integrable with respect to $\varepsilon\sim\mathcal P$. Then:
\begin{enumerate}
    \item $\psi(z)$ is convex in $z$, and the optimization problem is a convex
    optimization problem;

    \item let $\eta^\ast(z)$ be an optimal threshold in the
    Rockafellar--Uryasev representation of $\mathrm{CVaR}_{\beta}$. For each
    realization $\varepsilon$, let
    $\xi_{\eta^\ast(z)}(\varepsilon)\in[0,1]$ satisfy $\xi_{\eta^\ast(z)}(\varepsilon) \in \partial_u \bigl(u-\eta^\ast(z)\bigr)_+\Big|_{u=\ell_\alpha(f_0+\varepsilon,g(z))}$ and $\mathbb E_{\varepsilon\sim\mathcal P}\left[\xi_{\eta^\ast(z)}(\varepsilon)\right]=1-\beta$. 
    Then a subgradient of $\psi$ at $z$ is
    $$
    \frac{\alpha}{1-\beta}\mathbb E_{\varepsilon\sim\mathcal P}
    \left[\xi_{\eta^\ast(z)}(\varepsilon)V^\top \left(w(f_0+\varepsilon)-w\bigl(\alpha g(z)-f_0-\varepsilon\bigr)\right)\right]\in\partial\psi(z).
    $$
\end{enumerate}
\end{theorem}

\proof{Proof.}
For a fixed realization $\varepsilon$, define
$$
Y_z(\varepsilon):=\ell_{\alpha}(f_0+\varepsilon,g(z)).
$$
By Proposition~\ref{prop:spo_plus_ours},
$\ell_{\alpha}(f_0+\varepsilon,\cdot)$ is convex. Since
$g(z)=\bar f+Vz$ is affine, $Y_z(\varepsilon)$ is convex in $z$.

By the Rockafellar-Uryasev representation,
$$
\psi(z)=\min_{\eta\in\mathbb R}\left\{\eta+\frac{1}{1-\beta}
\mathbb E_{\varepsilon\sim\mathcal P}\left[\bigl(Y_z(\varepsilon)-\eta\bigr)_+\right]\right\}.
$$
Define
$$
\varphi(z,\eta):=\eta+\frac{1}{1-\beta}\mathbb E_{\varepsilon\sim\mathcal P}\left[\bigl(Y_z(\varepsilon)-\eta\bigr)_+\right].
$$
For each realization $\varepsilon$,
$(Y_z(\varepsilon)-\eta)_+$ is jointly convex in $(z,\eta)$ because it is
the maximum of the convex function $Y_z(\varepsilon)-\eta$ and zero.
Expectation preserves convexity, and hence $\varphi(z,\eta)$ is jointly convex
in $(z,\eta)$. Therefore, partial minimization over $\eta$ preserves convexity, so $\psi(z)=\min_{\eta\in\mathbb R}\varphi(z,\eta)$ $\{z:z^\top Hz\le r^2\}$ is convex. Thus, the CVaR formulation is a convex
optimization problem.

We next establish the subgradient characterization. By
Proposition~\ref{prop:spo_plus_ours}, for each realization $\varepsilon$,
$$
\alpha\left[w(f_0+\varepsilon)-w\bigl(\alpha g(z)-f_0-\varepsilon\bigr)\right]\in \partial_f \ell_{\alpha}(f_0+\varepsilon,g(z)).
$$
Since $g(z)=\bar f+Vz$ is affine,
$$
\alpha V^\top \left[w(f_0+\varepsilon)-w\bigl(\alpha g(z)-f_0-\varepsilon\bigr)\right]\in \partial_z Y_z(\varepsilon).
$$

Now fix an optimal threshold
$\eta^\ast(z)\in\arg\min_{\eta}\varphi(z,\eta)$. For each realization,
choose
$$
\xi_{\eta^\ast(z)}(\varepsilon) \in \partial_u \bigl(u-\eta^\ast(z)\bigr)_+\Big|_{u=Y_z(\varepsilon)},
$$
where
$$
\xi_{\eta^\ast(z)}(\varepsilon)=
\begin{cases}
0, & Y_z(\varepsilon)<\eta^\ast(z),\\
[0,1], & Y_z(\varepsilon)=\eta^\ast(z),\\
1, & Y_z(\varepsilon)>\eta^\ast(z).
\end{cases}
$$
The optimality of $\eta^\ast(z)$ implies that the selection at the threshold
can be chosen so that
$$
\mathbb E_{\varepsilon\sim\mathcal P}\left[\xi_{\eta^\ast(z)}(\varepsilon)\right]=1-\beta.
$$
Applying the subgradient chain rule and taking expectations therefore gives
$$
\frac{\alpha}{1-\beta}\mathbb E_{\varepsilon\sim\mathcal P}
\left[\xi_{\eta^\ast(z)}(\varepsilon)V^\top \left(w(f_0+\varepsilon)-w\bigl(\alpha g(z)-f_0-\varepsilon\bigr)
\right)\right]\in \partial\psi(z),
$$
which proves the stated subgradient expression.
\hfill$\square$
\endproof

\section{Demonstration Decision Problem}\label{app:decision_problem}
We consider a stylized investment allocation problem faced by an urban transportation authority. The authority is endowed with a fixed capital budget $B$ and seeks to allocate this budget across a set of candidate road segments in order to reduce aggregate commuter travel time in the city. Typical interventions include road widening, signal timing optimization, and intersection redesign, each of which improves link-level travel efficiency.

Let $J = \{1, \dots, m\}$ denote the set of candidate road segments eligible for investment. Prior to improvement, segment $j \in J$ has a baseline per-traveler travel time $t_j^0$. The city’s travel demand is represented by OD demand vector $f \in \mathbb{R}_+^n$, where $f_i$ denotes the daily travel demand associated with OD pair $i \in N := \{1, \dots, n\}$. The demand vector $f$ is generated by an upstream prediction model and serves as the key informational input to the downstream investment allocation decision. Network connectivity between OD pairs and road segments is described by a binary incidence matrix $S \in \{0,1\}^{n \times m}$, 
where $S_{ij} = 1$ if OD pair $i$ traverses segment $j$, and $S_{ij} = 0$ otherwise. We abstract from endogenous route choice and network equilibrium effects. Each OD pair is assumed to follow a representative baseline path, and link-level improvements are aggregated along this fixed path to obtain OD-level travel time. For each road segment $j \in J$, let $x_j$ denote the expected reduction (in minutes) in per-traveler travel time achieved through investment on that segment. We impose an upper bound $u_j$ reflecting engineering or physical feasibility constraints, here $u_j \le t_j^0$. Let $w_i$ denote the per trip travel time for OD pair $i$ as a result of the investments. Let $c_j$ denote the marginal cost of reducing one unit (minute) of travel time on segment $j$. Thus, the total investment cost on segment $j$ equals $c_j x_j$, and the aggregate expenditure across all segments must not exceed the available budget $B$. The sets, parameters, and decision variables are summarized in Table \ref{tab:notation}

\begin{table}[htbp]
\centering
\caption{Notation Summary for the Demonstration Decision Problem}
\label{tab:notation}
\begin{tabular}{ll}
\hline
\multicolumn{2}{l}{\textbf{Sets}}                                                               \\ \hline
$J = \{1,\dots,m\}$ & Set of candidate road segments eligible for investment                    \\
$N = \{1,\dots,n\}$ & Set of OD pairs                                     \\ \hline
\multicolumn{2}{l}{\textbf{Parameters}}                                                         \\ \hline
$f_i$               & Daily travel demand associated with OD pair $i$                           \\
$S_{ij}$            & Binary incidence indicator; $S_{ij}=1$ if OD pair $i$ traverses segment $j$     \\
$c_j$               & Marginal cost of reducing one unit (minute) of travel time on segment $j$ \\
$u_j$               & Maximum feasible travel time reduction on segment $j$                     \\
$B$                 & Total available investment budget                                         \\ \hline
\multicolumn{2}{l}{\textbf{Decision Variables}}                                                 \\ \hline
$x_j$               & Per-trip travel time reduction (minutes) achieved on segment $j$          \\
$w_i$               & Per-trip travel time (minutes) for OD pair $i$                  \\ \hline
\end{tabular}
\end{table}

The transportation authority seeks to minimize the total system-wide travel time experienced by all commuters. Because $f_i$ represents the demand for OD pair $i$, the aggregate daily travel time equals $\sum_{i \in N} f_i w_i$. The resulting linear program is:

\begin{align}
\min_{x,w} \quad 
& \sum_{i \in N} f_i w_i 
\label{eq:objective} \\[6pt]
\text{s.t.} \quad 
& w_i = \sum_{j \in J} S_{ij} (t_j^0-x_j), 
&& \forall i \in N, 
\label{eq:od_link} \\[6pt]
& \sum_{j \in J} c_j x_j \le B, 
\label{eq:budget} \\[6pt]
& 0 \le x_j \le u_j, 
&& \forall j \in J. 
\label{eq:bounds}
\end{align}

Constraint \eqref{eq:od_link} defines the post investment travel time for each OD pair as the sum of the improved link travel times along its route. Constraint \eqref{eq:budget} enforces the budget limitation. Constraint \eqref{eq:bounds} ensures feasibility of link-level improvements. This formulation captures a fundamental trade-off in transportation planning: investments on heavily utilized corridors yield large aggregate benefits, yet marginal improvement costs and physical constraints limit achievable reductions.

\section{Supplementary Details for Experiment \ref{EX2}}
\label{app:od_generation}

\subsection{Feature and OD Flow Generation}
For each experimental instance, we generate synthetic OD-flow samples from a
nonlinear feature-to-demand mapping. Each sample is associated with a feature
vector \(x_i\in\mathbb{R}^{d_{\mathrm{feat}}}\), where \(d_{\mathrm{feat}}=4\).
The first three feature dimensions are independently drawn from a standard
normal distribution, while the fourth dimension is a binary indicator:
$$
x_{i1},x_{i2},x_{i3}\sim \mathcal{N}(0,1),
\qquad
x_{i4}\sim \mathrm{Bernoulli}(0.5).
$$
Training and testing feature vectors are generated independently using different
random seeds.

Let \(n_{\mathrm{OD}}\) denote the number of OD pairs. For a given polynomial
order \(P\in\{1,\ldots,6\}\), the latent OD-flow vector is generated by
$$
Z_i=b+x_i W_1^\top+\sum_{p=2}^{P}x_i^{\circ p} W_p^\top,
$$
where \(x_i^{\circ p}\) denotes the elementwise \(p\)-th power of the feature
vector.

The nonlinear mapping is parameterized so that lower-order terms capture the
dominant feature effects, while higher-order terms introduce progressively
smaller nonlinear perturbations. Specifically, for each experimental instance, we first draw an intercept vector \(b\in\mathbb{R}^{n_{\mathrm{OD}}}\) and a
collection of coefficient matrices
\(\{W_p\}_{p=1}^{P}\), where \(W_p\in\mathbb{R}^{n_{\mathrm{OD}}\times d_{\mathrm{feat}}}\).
The entries are sampled independently according to
$$
b_j \sim U(0,3),\qquad (W_p)_{jk}\sim U(-\rho_p,\rho_p),
$$
with
$$
(\rho_1,\rho_2,\rho_3,\rho_4,\rho_5,\rho_6)=(1,\;0.5,\;0.2,\;0.1,\;0.05,\;0.02).
$$
Only the matrices \(W_1,\ldots,W_P\) are used for a polynomial order \(P\).
This scaling prevents high-order polynomial terms from dominating the generated
OD flows, while still allowing the true demand function to become increasingly
nonlinear as \(P\) increases.

Given the latent value \(Z_i\), the clean OD-flow vector is constructed through
a softplus transformation, $\widetilde Q_i=\log(1+\exp(Z_i))$, which ensures nonnegativity. The transformed values are then rescaled to the
target OD-flow magnitude and truncated to a bounded interval:
$$
Q_i=\Pi_{[0,2000]}\left(s\,\widetilde Q_i\right),
$$
where \(s>0\) is a fixed scaling constant and
\(\Pi_{[0,2000]}(\cdot)\) denotes componentwise projection onto the interval
\([0,2000]\). This step keeps the generated OD flows on a stable numerical
scale and avoids unrealistically large demands induced by rare high-order
polynomial values. The scaling factor places the OD flows on a realistic numerical scale for the
downstream investment problem, while the clipping step avoids unrealistically
large demands caused by rare high-order polynomial values.

\section{Supplementary Details for Experiment \ref{EX3}}

\subsection{OD Demand Patterns Generation}\label{app:od_patterns}

\paragraph{Common construction.}
Let $m\in\{1,\ldots,5\}$ index the five demand-generation modes. For each OD pair $(i,j)$, let $d_{ij}$ denote the shortest-path distance,
normalized by the median of all positive shortest-path distances in the network. For each mode $m$, we first construct a base OD pattern
$$
\widetilde F^{(m)}=(1-\lambda)B+\lambda S^{(m)},\qquad \lambda=0.85,
$$
where $B$ is a common nonnegative gravity-type background and $S^{(m)}$ is the nonnegative mode-specific component.

The common background $B$ incorporates heterogeneous origin-production
and destination-attraction effects, moderate distance decay, and mild
cell-level variation. It introduces a shared level of spatial heterogeneity
across all modes and prevents the resulting OD patterns from becoming
artificially sparse. The larger weight on $S^{(m)}$ preserves the
distinctive structure of each demand-generation mode. Intrazonal demand
is set to zero.

\paragraph{Pattern variants.}
As illustrated in Figure~\ref{fig:siouxfalls-patterns}, the five
mode-specific components $S^{(m)}$ are constructed as follows.

\begin{enumerate}
    \item \textit{Distance decay.}
    Demand is generated from heterogeneous origin and destination
    propensities and decreases exponentially with network distance.
    This pattern represents a conventional gravity-type structure in
    which nearby OD pairs interact more strongly.

    \item \textit{Low-rank latent demand.}
    The demand signal is obtained from the nonnegative part of a
    rank-three bilinear factor model and is moderately attenuated by
    distance. This construction captures correlated OD movements driven
    by a small number of latent factors.

    \item \textit{Community blocks.}
    Nodes are randomly assigned to three communities, with substantially
    stronger demand within communities than across communities.
    Cell-level variation is retained so that flows remain heterogeneous
    within each block.

    \item \textit{Directional corridors.}
    Disjoint source and sink sets, each containing approximately one
    quarter of the nodes, are selected to form dominant source-to-sink
    corridors. Corridor demand decreases with distance, while a weak
    diffuse component allows residual demand outside the primary
    corridors, producing a strongly asymmetric OD structure.

    \item \textit{Hub--spoke concentration.}
    The node with the highest network degree is designated as the hub.
    Demand is concentrated on trips originating from or terminating at
    the hub, with a small residual component representing interactions
    among nonhub nodes.
\end{enumerate}

\paragraph{Demand scenarios.}
Let $\ell\in\{1,2,3\}$ index the low, medium, and high demand levels,
with prescribed aggregate demands $Q_\ell\in\{1{,}000,\;10{,}000,\;20{,}000\}$, respectively. Combining the five demand-generation modes with the three demand levels yields 15 scenarios.

To ensure that the three levels are not simple scalar copies of the same OD pattern, we introduce level-specific origin and destination effects.
For each level $\ell$, define
$$
M_{ij}^{(\ell)} = \exp\left(\alpha_i^{(\ell)}+\beta_j^{(\ell)}\right),
$$
where $\alpha_i^{(\ell)}$ and $\beta_j^{(\ell)}$ denote the origin- and destination-specific effects, respectively. The effects are centered, while their dispersion varies across demand levels.

The ground-truth OD matrix for mode $m$ and demand level $\ell$ is then
constructed as
$$
F_{\mathrm{true}}^{(m,\ell)}=\operatorname{Scale}_{Q_\ell}\left(\widetilde F^{(m)}\odot M^{(\ell)}\right),
$$
where $\odot$ denotes elementwise multiplication and, for any nonnegative matrix $A$ with zero diagonal,
$$
\operatorname{Scale}_{Q}(A)=\frac{Q}{\sum_{i\neq j}A_{ij}}A.
$$
Hence, $\sum_{i\neq j}F_{\mathrm{true},ij}^{(m,\ell)}=Q_\ell$.

For each demand level $\ell$, the same multiplier $M^{(\ell)}$ is applied across all five modes, and fixed random seeds are used throughout. This design preserves comparability across modes while allowing both
aggregate demand and OD margins to vary across demand levels.

Finally, each ground-truth OD matrix is vectorized as
$f_{\mathrm{true}}^{(m,\ell)}$ and mapped to link flows through the common assignment relation
$$
L^{(m,\ell)}=X f_{\mathrm{true}}^{(m,\ell)}.
$$
Thus, all scenarios share the same network topology and assignment mechanism, while varying only in their OD structure and demand level.

\subsection{Repeated-Sampling Robustness Analysis}
\label{app:repeated_sampling}

To examine whether the conclusions from the Sioux Falls experiment are robust to variation across experimental realizations, we conduct an additional repeated-sampling analysis. Each of the 15 demand-level--mode combinations is independently evaluated over 20 experimental realizations generated using distinct random seeds. Table~\ref{tab:robust} reports the sample mean and sample standard deviation of decision regret across these realizations.

\begin{table}[htbp]
\centering
\caption{Repeated-sampling decision regret across demand settings on the Sioux Falls network.}
\begin{threeparttable}

\label{tab:robust}
\small
\setlength{\tabcolsep}{4pt}
\renewcommand{\arraystretch}{1.15}

\begin{tabular}{llrrrrr}
\toprule
\multirow{2}{*}{\textbf{Demand Level}}
& \multirow{2}{*}{\textbf{Demand Mode}}
& \multicolumn{4}{c}{\textbf{Mean Decision Regret (SD)}}
& \multirow{2}{*}{$\boldsymbol{r^\ast}$} \\
\cmidrule(lr){3-6}
&
& \textbf{OLS}
& \textbf{PTO}
& \textbf{D-FULL-G}
& \textbf{D-GEO-G}
& \\
\midrule

\multirow{5}{*}{Low}
& Distance Decay
& 18.51 (7.08)
& 3.62 (2.69)
& 1.37 (1.27)
& \textbf{1.29} (1.23)
& 5 \\

& Low Rank
& 51.58 (23.68)
& 21.95 (20.49)
& \textbf{4.09} (5.35)
& \textbf{4.09} (5.35)
& 10 \\

& Community Blocks
& 53.70 (18.13)
& 29.06 (21.74)
& \textbf{3.71} (3.27)
& \textbf{3.71} (3.27)
& 10 \\

& Directional Corridor
& 48.65 (28.08)
& 11.79 (10.98)
& \textbf{4.59} (5.56)
& 5.03 (5.48)
& 10 \\

& Hub Spoke
& 107.04 (28.83)
& 5.12 (4.37)
& \textbf{1.69} (1.94)
& 2.27 (2.65)
& 10 \\

\midrule

\multirow{5}{*}{Medium}
& Distance Decay
& 211.90 (113.20)
& 33.70 (24.28)
& 14.92 (15.19)
& \textbf{12.04} (13.62)
& 3 \\

& Low Rank
& 531.08 (235.01)
& 235.79 (163.89)
& \textbf{34.39} (57.87)
& 34.98 (58.72)
& 10 \\

& Community Blocks
& 596.87 (180.51)
& 282.17 (224.14)
& \textbf{32.04} (35.70)
& \textbf{32.04} (35.70)
& 10 \\

& Directional Corridor
& 468.94 (320.83)
& 119.44 (143.96)
& \textbf{53.01} (69.35)
& 59.26 (78.96)
& 10 \\

& Hub Spoke
& 1098.04 (344.61)
& 57.66 (42.63)
& 17.39 (15.50)
& \textbf{16.25} (15.10)
& 10 \\

\midrule

\multirow{5}{*}{High}
& Distance Decay
& 345.48 (190.05)
& 65.71 (47.21)
& 23.94 (22.17)
& \textbf{22.58} (21.82)
& 5 \\

& Low Rank
& 984.47 (413.93)
& 490.46 (361.87)
& \textbf{72.98} (61.36)
& \textbf{72.98} (61.36)
& 10 \\

& Community Blocks
& 1184.09 (409.70)
& 619.09 (430.08)
& \textbf{98.47} (78.00)
& \textbf{98.47} (78.00)
& 10 \\

& Directional Corridor
& 1059.13 (518.97)
& 277.58 (229.15)
& 92.71 (88.28)
& \textbf{88.32} (89.99)
& 10 \\

& Hub Spoke
& 1924.63 (651.64)
& 106.03 (91.86)
& \textbf{43.43} (45.71)
& 44.43 (47.86)
& 10 \\

\bottomrule
\end{tabular}

\begin{tablenotes}[flushleft]
\footnotesize
\item[] \textit{Notes.}
Values are reported as mean (sample standard deviation (SD)) over 20
independent experimental realizations.
All decision-regret values are divided by $10^3$. For D-GEO-G, $r^\ast$ denotes the radius attaining the lowest mean decision regret among the tested radii. Bold values indicate the lowest mean decision regret within each demand setting, with both values highlighted in the case of a tie.
\end{tablenotes}

\end{threeparttable}
\end{table}

The repeated-sampling results provide consistent evidence for the advantage of decision aware estimation. In every one of the 15 demand-level--mode combinations, both D-FULL-G and D-GEO-G achieve strictly lower mean decision regret than PTO, with no exception across the three demand levels or the five demand modes. Relative to PTO, the reduction in mean regret ranges from $55.6\%$ to $88.6\%$ for D-FULL-G and from $50.4\%$ to $88.6\%$ for D-GEO-G. Averaged across all settings, mean regret decreases from $157.28$ under PTO to $33.25$ under D-FULL-G and $33.18$ under D-GEO-G, corresponding to an average reduction of approximately $78.9\%$ for either variant. The improvement is also reflected in the variability across experimental realizations: both DWE variants have a smaller sample standard deviation than PTO in all 15 settings. For example, under high-demand Community Blocks, PTO produces a mean regret of $619.09$ with a standard deviation of $430.08$, whereas both D-FULL-G and D-GEO-G reduce these values to $98.47$ and $78.00$, respectively. Similarly, under medium-demand Directional Corridor, D-FULL-G and D-GEO-G reduce the PTO mean regret from $119.44$ to $53.01$ and $59.26$, while their standard deviations decrease from $143.96$ to $69.35$ and $78.96$. Thus, the gains from decision aware estimation are not limited to lower average regret; they are accompanied by consistently lower variation across repeated realizations. The comparison between the two DWE variants does not indicate uniform dominance. D-GEO-G achieves lower mean regret than D-FULL-G in five settings, matches it in five settings, and has higher mean regret in the remaining five, while their overall averages remain nearly identical. Its largest relative improvement occurs under medium-demand Distance Decay, where mean regret decreases from $14.92$ to $12.04$, a reduction of $19.3\%$ relative to D-FULL-G. Its largest absolute shortfall occurs under medium-demand Directional Corridor, where mean regret increases from $53.01$ to $59.26$; nevertheless, D-GEO-G still achieves a strictly lower mean regret than PTO by $50.4\%$ in this setting. These differences reflect the trade-off introduced by the geometric restriction. D-FULL-G permits a more flexible adjustment, whereas D-GEO-G restricts the admissible adjustment through the radius $r$. This restriction can limit sample-specific over-adjustment, but it may also exclude corrections that are important to the downstream decision and thereby introduce approximation error. Notably, the selected radius reaches $r^\ast=10$, the maximum radius considered, in all five settings where D-GEO-G has higher mean regret than D-FULL-G. This boundary behavior is consistent with the admissible geometric region remaining too restrictive in these cases. The underperformance is concentrated in the Low Rank, Directional Corridor, and Hub Spoke modes, whose global, directional, or hub-centered structures may be less well represented by a finite geometric neighborhood. Although this observation does not establish that increasing $r_{\max}$ would necessarily eliminate the performance gap, it identifies the finite geometric restriction as a plausible source of approximation error. D-GEO-G should therefore be interpreted as a regularized alternative to D-FULL-G rather than as a uniformly dominating variant. Importantly, its occasional disadvantage relative to D-FULL-G does not affect the central robustness conclusion: D-GEO-G, like D-FULL-G, achieves strictly lower mean regret and lower cross-realization variability than PTO in every tested setting.

\end{APPENDICES}

\end{document}